\documentclass[journal]{IEEEtran}

\usepackage{ntheorem}
\theoremheaderfont{\rmfamily\bfseries\upshape}
\theorembodyfont{\rmfamily\mdseries\itshape}
\theoremseparator{.}
\newtheorem{defi}{Definition}
\newtheorem{thm}{Theorem}
\newtheorem{lem}{Lemma}
\newtheorem{lems}{Lemma}
\newtheorem{coro}{Corollary}

\usepackage{bbm}
\usepackage{upgreek}
\usepackage{amsmath}
\usepackage{xcolor}
\usepackage{graphicx}
\usepackage{booktabs}
\usepackage{amssymb}
\usepackage{textcomp}
\usepackage{amsmath,bm}
\usepackage{makecell}
\usepackage{multirow}
\usepackage{caption}
\usepackage{float}
\usepackage{colortbl}
\usepackage{mathrsfs}
\usepackage{changepage}
\usepackage[colorlinks]{hyperref}
\usepackage{color}
\usepackage{soul}
\usepackage{tcolorbox}
\usepackage{subfigure}
\usepackage{tikz}
\usepackage{lipsum}
\usepackage[mathlines,switch]{lineno}
\usetikzlibrary{calc,tikzmark}
\usepackage[numbers,sort&compress]{natbib}
\makeatletter  
\newif\if@restonecol  
\makeatother

\usepackage[linesnumbered,ruled,vlined]{algorithm2e}
\usepackage{algpseudocode}
\usepackage{amsmath}  

\definecolor{myboxcolor}{gray}{0.97}
\definecolor{myred}{rgb}{1, 0, 0}

\usepackage{tikz,xcolor,hyperref}
\definecolor{lime}{HTML}{A6CE39}
\DeclareRobustCommand{\orcidicon}{%
    \begin{tikzpicture}
    \draw[lime, fill=lime] (0,0) 
    circle [radius=0.16] 
    node[white] {{\fontfamily{qag}\selectfont \tiny ID}};    \draw[white, fill=white] (-0.0625,0.095)
    circle [radius=0.007];    \end{tikzpicture}
    \hspace{-2mm}}
\foreach \x in {A, ..., Z}{%
    \expandafter\xdef\csname orcid\x\endcsname{\noexpand\href{https://orcid.org/\csname orcidauthor\x\endcsname}{\noexpand\orcidicon}}
    }
\hypersetup{hidelinks}

\ifCLASSINFOpdf

\else

\fi

\begin{document}
\IEEEoverridecommandlockouts
\IEEEpubid{\begin{minipage}[t]{\textwidth}\ \\[0.6pt]
        \centering\footnotesize{\begin{tcolorbox}[left = 0.5mm, right = 0.5mm, top = 0.5mm, bottom = 0.5mm]\copyright This work has been submitted to the IEEE for possible publication. Copyright may be transferred without notice, after which this version may no longer be accessible.\end{tcolorbox}}
\end{minipage}}

\title{Surrogate-Assisted Search with Competitive Knowledge Transfer for Expensive Optimization}

\author{Xiaoming Xue\orcidA{},
		Yao Hu\orcidB{}, \IEEEmembership{~Student Member,~IEEE},
		Liang Feng\orcidD{}, \IEEEmembership{~Senior Member,~IEEE},\par
		Kai Zhang\orcidF{}, \IEEEmembership{~Member,~IEEE},
		Linqi Song\orcidG{},~\IEEEmembership{~Senior Member,~IEEE},
		and Kay Chen Tan\orcidH{},~\IEEEmembership{~Fellow,~IEEE}
\thanks{Xiaoming Xue, Yao Hu, and Linqi Song are with the Department of Computer Science, City University of Hong Kong, Hong Kong SAR, China and also with the City University of Hong Kong Shenzhen Research Institute, Shenzhen 518057, China (e-mail: xminghsueh@gmail.com; y.hu@my.cityu.edu.hk; linqi.song@cityu.edu.hk).}
\thanks{Liang Feng is with the College of Computer Science, Chongqing University, Chongqing 400044, China (e-mail: liangf@cqu.edu.cn).}
\thanks{Kai Zhang is with the Civil Engineering School, Qingdao University of Technology, Qingdao 266520, China (e-mail: zhangkai@qut.edu.cn).}
\thanks{Kay Chen Tan is with the Department of Data Science and Artificial Intelligence, The Hong Kong Polytechnic University, Hong Kong SAR, China (e-mail: kctan@polyu.edu.hk).}
}


\maketitle

\begin{abstract}
Expensive optimization problems (EOPs) have attracted increasing research attention over the decades due to their ubiquity in a variety of practical applications.
Despite many sophisticated surrogate-assisted evolutionary algorithms (SAEAs) that have been developed for solving such problems, most of them lack the ability to transfer knowledge from previously-solved tasks and always start their search from scratch, making them troubled by the notorious cold-start issue.
A few preliminary studies that integrate transfer learning into SAEAs still face some issues, such as defective similarity quantification that is prone to underestimate promising knowledge, surrogate-dependency that makes the transfer methods not coherent with the state-of-the-art in SAEAs, etc.
In light of the above, a plug and play competitive knowledge transfer method is proposed to boost various SAEAs in this paper.
Specifically, both the optimized solutions from the source tasks and the promising solutions acquired by the target surrogate are treated as task-solving knowledge, enabling them to compete with each other to elect the winner for expensive evaluation, thus boosting the search speed on the target task.
Moreover, the lower bound of the convergence gain brought by the knowledge competition is mathematically analyzed, which is expected to strengthen the theoretical foundation of sequential transfer optimization.
Experimental studies conducted on a series of benchmark problems and a practical application from the petroleum industry verify the efficacy of the proposed method.
The source code of the competitive knowledge transfer is available at \textcolor{magenta}{\url{https://github.com/XmingHsueh/SAS-CKT}}.
\end{abstract}


\IEEEpeerreviewmaketitle


\section{Introduction}
\label{section:intro}

Expensive optimization problems (EOPs)~\cite{ong2003evolutionary,jin2018data,li2022evolutionary} refer to the problems whose objective functions or constraints involve expensive or even unaffordable evaluations, which widely exist in many real-world applications.
Hyperparameter tuning of deep learning models~\cite{shankar2020hyperparameter}, high-fidelity numerical simulation~\cite{islam2014simulation}, and physical experiment-based optimization~\cite{yoruklu2022optimization} are just a few representative examples.
Despite the popularity of evolutionary algorithms (EAs) in solving various optimization problems due to their superior global search capabilities and domain information-agnostic implementations~\cite{liu2023survey}, they typically assume that evaluating the objectives or constraints of candidate solutions is not expensive and always require a large number of evaluations to obtain satisfactory solutions, making them computationally unaffordable for many EOPs.

Over the past decades, a variety of techniques have been developed to improve EAs on EOPs, which can be divided into three categories according to the cost reduction mechanism in~\cite{li2022evolutionary}: 1) problem approximation and substitution~\cite{lim2009generalizing}, 2) algorithm enhancement~\cite{zychowski2018addressing}, and 3) parallel and distributed computing.
In particular, surrogate-assisted search (SAS) in the first category has been gaining great popularity due to its ease of implementation and good generality~\cite{he2023review}.
Polynomial response surface~\cite{box1978statistics}, Gaussian process regression~\cite{williams2006gaussian} and radial basis function~\cite{hardy1971multiquadric} are a few representative surrogates.
Over the years, a growing number of surrogate-assisted evolutionary algorithms (SAEAs) have been proposed to combat different complexities of EOPs, including objective conflicts in multiobjective problems~\cite{zhang2009expensive}, high-dimensional variables in large-scale optimization~\cite{sun2022surrogate}, and robust optimization for problems with uncertainties~\cite{wang2019surrogate}, to name a few.
However, despite numerous promising results reported, most SAEAs lack the capability to transfer knowledge from potentially similar tasks towards enhanced problem-solving efficiency.
Oftentimes, they consider potentially related tasks in isolation and start the search of each individual task from scratch~\cite{Wang2023RecentBO}, which often require a considerable number of expensive evaluations to achieve high-quality solutions.
This phenomenon is also known as the cold-start issue in the literature~\cite{zhu2023federated}.

To alleviate the cold-start issue, a growing number of studies propose to endow SAEAs with the ability to transfer knowledge from possibly related tasks for achieving better performance~\cite{tan2021evolutionary}.
According to the availability of source tasks, knowledge transfer can be divided into three categories~\cite{gupta2017insights}: sequential transfer~\cite{feng2017autoencoding,shakeri2022scalable,liu2024extremo}, multitasking~\cite{gupta2015multifactorial,liu2022evolutionary,wang2024evolutionary}, and multiform optimization~\cite{guo2022generative,feng2023multiform}.
In cases where no source tasks are available, one can generate alternate formulations of a target task of interest and employ them as helper tasks to better solve the target task~\cite{yang2020offline,feng2021multivariation}, which is known as multiform optimization.
By contrast, if a certain number of unoptimized tasks that are potentially related to the target task are available, one can solve them with the target task in a multitasking manner to exploit inter-task synergies for improved search performance~\cite{ji2023multisurrogate}.
Furthermore, when a number of previously-solved tasks are available as the source tasks, one can extract transferable cues from them to guide the optimization of the target task~\cite{min2019multiproblem,min2021generalizing}, which is termed sequential transfer.

In this study, we focus on improving SAEAs with sequential transfer, due to its widespread application scenarios in the realm of engineering design, where many problems necessitate frequent re-optimization in view of changing environments or problem features, such as aerodynamic shape design with new conditions and constraints~\cite{yan2019aerodynamic}, well placement optimization of newfound reservoirs~\cite{qi2023evolutionary}, geological parameter estimation of a series of core plugs from nearby blocks~\cite{li2022bayesian}, and heat extraction optimization of geothermal systems~\cite{chen2024surrogate}.
As time goes by, an increasing number of optimization tasks accumulate in a database, providing an opportunity for the target task at hand to achieve better optimization performance via knowledge transfer.
In the literature, some attempts have been made to improve SAEAs through sequential knowledge transfer.
For instance, in~\cite{min2019multiproblem}, sequential transfer is conducted by aggregating the source surrogate models with the target one, where the weights for capturing source-target similarities are estimated by minimizing the squared error of out-of-sample predictions.
However, the aggregation-based similarity is based on the objective values of solutions instead of their ranks, which tends to underestimate the similarity between two tasks with quantitatively different objective responses but similar ranks standing for high solution transferability~\cite{bardenet2013collaborative}.
Besides, the lack of task adaptation in the aggregation model makes it less effective in dealing with heterogeneous tasks.
To address this issue, a generalized transfer Bayesian optimization algorithm that employs neural networks to overcome the task heterogeneity is developed in~\cite{min2021generalizing}.
However, this algorithm shows poor portability due to its dependency on the Bayesian models and specific applicability to single-source problems.

With the above in mind, this paper proposes a plug and play knowledge transfer module for enhancing the problem-solving capability of SAEAs.
A novel mechanism named competitive knowledge transfer is proposed, which assesses the promising solutions obtained by SAS on the target task and the optimized solutions from the source tasks from a consistent view.
By treating both of these two types of solutions as task-solving knowledge for competition, it selects the winner to undergo the real function evaluation.
Overall, the proposed surrogate-assisted search with competitive knowledge transfer (SAS-CKT) exhibits three main merits: 1) portability that makes it convenient to be used for improving different SAEAs and thus coherent with the state-of-the-art; 2) reliability of similarity quantification that enables knowledge transfer across two tasks with high solution transferability regardless of their different orders of magnitude; 3) adaptivity that allows it to deal with the task heterogeneity.
Experimental studies conducted on a series of benchmark problems and a practical application demonstrate the efficacy of SAS-CKT and its superiority against a few state-of-the-art algorithms.
The main contributions of this paper are summarized as follows:

\begin{figure}[ht]
	\centering
	\includegraphics[width=2.7in]{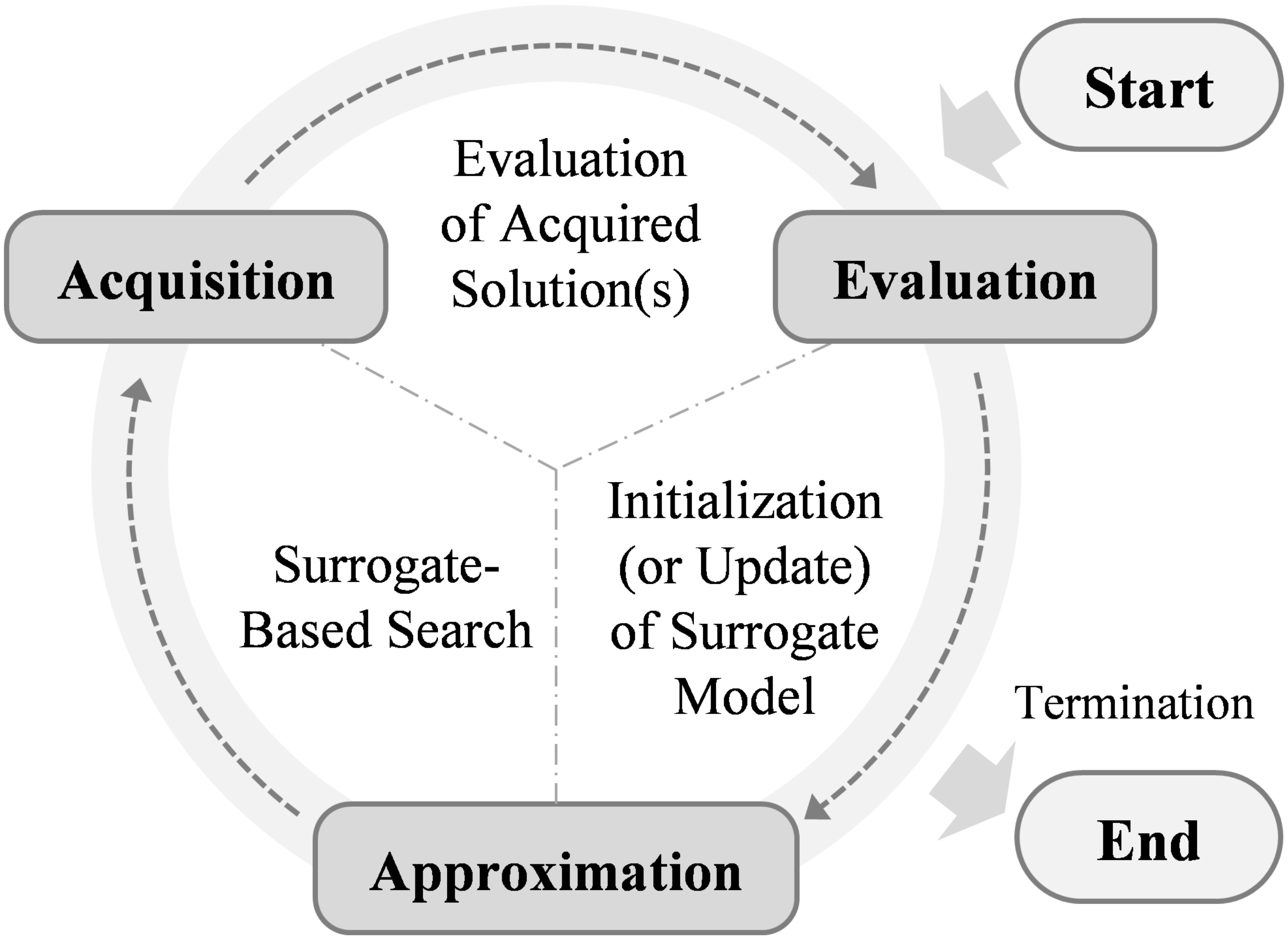}
	\caption{High-level structure of SAS.}
	\label{fig:saes}
\end{figure}

\begin{itemize}
\item A plug and play competitive knowledge transfer method is proposed to enable prompt performance improvement of different SAEAs.
Specifically, by treating both the source and target solutions as task-solving knowledge and assessing them from a consistent view, the winner is identified for expensive evaluation.
\item We conduct theoretical analyses to prove that the lower bound of the convergence gain brought by the proposed method is bounded by zero regardless of the source-target similarity.
Moreover, important conditions that could lead to positive convergence gain are analyzed in detail.
\item The efficacy of the proposed method is empirically validated on a series of benchmark problems, as well as a practical case study from the petroleum industry.
\end{itemize}

The remainder of this paper is organized as follows.
Section \ref{section:pre} presents the definitions of SAS and sequential transfer.
After that, a few preliminary studies that improve SAS with sequential transfer are reviewed and analyzed, followed by the motivations behind this work.
Then, Section \ref{section:ckt} introduces the proposed SAS-CKT in detail and Section \ref{section:the} presents its theoretical analyses.
With the experimental settings provided in Section \ref{section:setup}, we conduct our experiments and analyze the corresponding results in Section \ref{section:exp}.
Lastly, Section \ref{section:con} concludes this paper.


\section{Preliminaries}
\label{section:pre}

In this section, we first briefly introduce the canonical SAS paradigm and the sequential transfer-enhanced optimization.
After that, the limitations of a few preliminary studies that improve SAS with sequential transfer are analyzed.
Lastly, the motivations behind this work are presented.

\subsection{Surrogate-Assisted Search}

Surrogate-assisted search (SAS) is a widely used optimization paradigm for computationally expensive problems, which consists of the following three key parts~\cite{he2023review}:
\begin{itemize}
\item[1)] \emph{Approximation:} Construct (or update) a surrogate model to approximate the objective function of interest.
\item[2)] \emph{Acquisition:} Conduct the global or local search on the surrogate model with specific optimizers (e.g., evolutionary algorithms or nonlinear programming techniques) to acquire a promising solution.
\item[3)] \emph{Evaluation:} Evaluate the promising solution by calling the real function evaluation and update the database with the newly evaluated solution accordingly.
\end{itemize}

\begin{table*}[ht]
	\caption{A list of state-of-the-art surrogate-assisted search algorithms.}
	\centering
	\footnotesize
	\heavyrulewidth=0.12em
	\lightrulewidth=0.1em
	\cmidrulewidth=0.1em
	\setlength\tabcolsep{3.7pt}
	\begin{tabular}{lllllll}
		\toprule
\multirow{3}*{Algorithm}&\multirow{3}*{Year}&\multicolumn{2}{c}{Surrogate Model}&\multicolumn{2}{c}{Search Mechanism}&\multicolumn{1}{c}{\multirow{3}*{Infill Criterion}}\\
		\cmidrule(){3-4} \cmidrule(l){5-6}
		&&\makecell[c]{Type}&\makecell[c]{Model}&\makecell[c]{Prescreening or Iteration}&\makecell[c]{Optimizer}&\\
		\midrule
\makecell[l]{IKAEA~\cite{zhan2021fast}}&2021&Global&Gaussian process regression (GPR)&Prescreening&\makecell[l]{Differential\\evolution (DE)}&Expected improvement (EI)\\
		\midrule
		\multirow{4.5}*{TLRBF~\cite{li2021three}}&\multirow{4.5}*{2022}&Global&\makecell[c]{\multirow{4.5}*{Radial basis function (RBF)}}&Prescreening&Random sampling&\makecell[c]{\multirow{4.5}*{Predicted objective value (POV)}}\\
		\cmidrule(){3-3}\cmidrule(){5-6}
		&&Subregion&&\multirow{3}*{Iteration}&\multirow{3}*{\makecell[l]{Adaptive DE\\(JADE)}}&\\
		\cmidrule(){3-3}
		&&Local&&&&\\
		\midrule
		\multirow{3}*{\makecell[l]{GL-SADE\\\cite{wang2022surrogate}}}&\multirow{3}*{2022}&Global&\makecell[c]{RBF}&Iteration&\makecell[c]{\multirow{3}*{DE}}&\makecell[c]{POV}\\
		\cmidrule(){3-5}\cmidrule(){7-7}
		&&Local&\makecell[c]{GPR}&Prescreening&&Lower confidence bound (LCB)\\
		\midrule
		\multirow{4.5}*{\makecell[l]{DDEA\\-MESS\\\cite{yu2022data}}}&\multirow{4.5}*{2022}&Global&\makecell[c]{\multirow{4.5}*{RBF}}&Prescreening&\makecell[c]{\multirow{3}*{DE}}&\makecell[c]{\multirow{4.5}*{POV}}\\
		\cmidrule(){3-3}\cmidrule(){5-5}
		&&Local&&\multirow{3}*{Iteration}&&\\
		\cmidrule(){3-3}\cmidrule(){6-6}
		&&Region&&&Trust region search&\\
		\midrule
		\multirow{4.5}*{\makecell[l]{LSADE\\\cite{kuudela2023combining}}}&\multirow{4.5}*{2023}&Global&\makecell[c]{RBF}&\multirow{4.5}*{Prescreening}&\makecell[c]{\multirow{4.5}*{DE}}&\makecell[c]{\multirow{4.5}*{POV}}\\
		\cmidrule(){3-4}
		&&Region&Lipschitz surrogate&&&\\
		\cmidrule(){3-4}
		&&Local&\makecell[c]{RBF}&&&\\
		\midrule
		\multirow{6}*{\makecell[l]{AutoSAEA\\\cite{xie2023surrogate}}}&\multirow{6}*{2024}&\multirow{6}*{Local}&\makecell[c]{GPR}&Prescreening&\makecell[c]{\multirow{6}*{DE}}&\makecell[c]{\{LCB, EI\}}\\
		\cmidrule(){4-5}\cmidrule(){7-7}
		&&&\makecell[c]{RBF}&\multirow{3}*{\{Prescreening, Iteration\}}&&\makecell[c]{\multirow{3}*{POV}}\\
		\cmidrule(){4-4}
		&&&Polynomial response surface (PRS)&&&\\
		\cmidrule(){4-5}\cmidrule(){7-7}
		&&&K-nearest neighbor (KNN)&Prescreening&&\{L1-exploitation, L1-exploration\}\\
		\bottomrule
	\end{tabular}
	\label{tab:sas_methods}
\end{table*}

Fig. \ref{fig:saes} shows the high-level structure of SAS.
It starts with the evaluation of a number of uniformly distributed solutions for initializing a surrogate model, which can be generated by a random sampling technique (e.g., the Latin hypercube sampling, LHS).
Polynomial response surface~\cite{box1978statistics}, Gaussian process regression~\cite{williams2006gaussian} and radial basis function~\cite{hardy1971multiquadric} are a few commonly used surrogate models.
Then, with a specific acquisition function, also known as infill criterion, the global or local search with particular optimizers is performed on the surrogate model instead of the real function to acquire a promising solution.
Subsequently, the promising solution is evaluated by the real function and added into the database for updating the surrogate model.
The above three phases are repeatedly executed until the termination condition is met, e.g., the maximum number of expensive evaluations is reached.

Despite a wide variety of SAS algorithms that have been developed over the decades, the majority of them follow the high-level structure shown in Fig. \ref{fig:saes}.
To demonstrate, we select a number of recently published SAS algorithms from different peer-reviewed journals and compare their components in Table \ref{tab:sas_methods}.
It can be seen that the advances in SAS mainly come from three aspects: 1) appropriate surrogate models for approximating the problem of interest; 2) novel search mechanisms for optimizing the surrogates; and 3) effective infill criteria for acquiring promising solutions.
Most recently, a sophisticated SAS framework with the auto-configuration of surrogate model and infill criterion is developed in~\cite{xie2023surrogate}, whose superiority against many other state-of-the-art SAEAs is validated on a number of complex benchmark problems and practical applications.
For more detailed introductions of SAS and comprehensive reviews of various SAS algorithms, interested readers may refer to~\cite{jin2018data,li2022evolutionary,he2023review}.

\subsection{Sequential Transfer}

With a number of previously-optimized source tasks stored in a knowledge base $\mathcal{M}$, a target task of interest is supposed to achieve better task-solving efficiency by transferring the knowledge extracted from $\mathcal{M}$~\cite{gupta2017insights}, as can be formulated by
\begin{equation}
Q\left(\mathcal{T}\mid\mathcal{M}\right)-Q\left(\mathcal{T}\right)>0,
\label{eq:sto_ps}
\end{equation}
where $\mathcal{T}$ is the target task, $Q\left(\mathcal{T}\right)$ indicates the efficiency of an optimizer without transfer, $Q\left(\mathcal{T}\mid\mathcal{M}\right)$ denotes the efficiency of the optimizer with the knowledge transferred from $\mathcal{M}$.

The sequential transfer in Eq. \eqref{eq:sto_ps} has been widely employed to improve the evolutionary search as a means of convergence speedup~\cite{Zhou2021Learnable,xue2022evolutionary,xue2023solution}.
Likewise, it provides a promising solution for alleviating the cold-start issue of SAS, which has fostered a number of relevant studies presented in what follows.

\subsection{Surrogate-Assisted Search Boosted by Sequential Transfer}
When solving black-box EOPs with SAS, the available information of the previously-optimized source tasks is simply their surrogate models with the associated evaluated solutions.
Given a target task to be optimized, its task-solving efficiency is expected to be improved with the knowledge extracted from the source data, as can be formulated by
\begin{equation}
\min_{\boldsymbol{x}\in\Omega}\left[f\left(\boldsymbol{x}\right)|\lbrace\hat{f}^s_i\left(\boldsymbol{x}\mid \mathcal{D}^s_i\right),\,1\le i\le k\rbrace\right],
\label{eq:sto_saes}
\end{equation}
where $\boldsymbol{x}$ denotes the decision vector, $\Omega$ is the decision space, $f\left(\boldsymbol{x}\right)$ is the objective function of the target task, $\hat{f}^s_i\left(\boldsymbol{x}\mid \mathcal{D}^s_i\right)$ denotes the surrogate model of the $i$-th source task built upon the source data $\mathcal{D}^s_i=\lbrack X_i,\boldsymbol{y}_i\rbrack$, $X_i$ and $\boldsymbol{y}_i$ denote the evaluated solutions and the associated objective values of the $i$-th source task, $k$ is the number of source tasks.

\begin{figure*}[ht]
	\centering
	\includegraphics[width=6.75in]{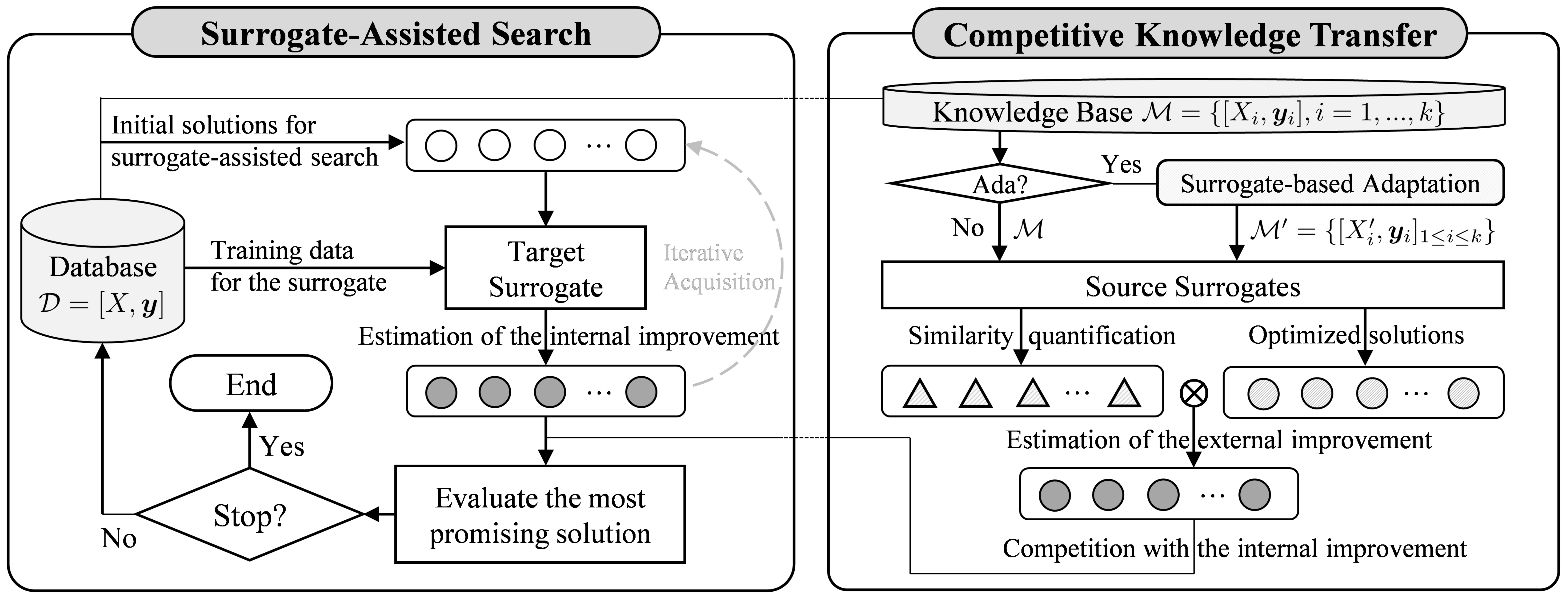}
	\caption{Flow chart of the proposed SAS-CKT.}
	\label{fig:flowchart}
\end{figure*}

Bayesian optimization is a representative SAS framework that uses Bayesian inferences (e.g., Gaussian process regression and Bayesian neural networks) as its surrogate models~\cite{shahriari2015taking}, which has been gaining increasing research efforts on endowing it with the ability to transfer knowledge from potentially related tasks for better optimization performance.
Prior incorporation~\cite{wang2018regret}, kernelization of cross-task samples~\cite{min2021generalizing} and ensemble inference~\cite{wistuba2016two} are three knowledge transfer techniques commonly used for improving Bayesian optimization.
However, all these methods are undetachably bonded to the associated Bayesian surrogate models, making them suffer from the downsides of those models, e.g., the poor scalability and the cubic complexity to the data size of Gaussian process~\cite{liu2020gaussian}.
Besides, the model-dependency of these knowledge transfer methods makes them disconnected from the advances made in general SAS methods.

In~\cite{min2019multiproblem}, a general ensemble inference model named multi-problem surrogates is proposed, in which the weights that capture source-target similarities are estimated by minimizing the squared error of out-of-sample predictions.
It is noted that this ensemble inference method is fairly general although it was instantiated by Gaussian process only in~\cite{min2019multiproblem}, enabling it to be used in conjunction with other SAS algorithms with ease.
However, for two optimization tasks with quantitatively different objective values but similar rank responses standing for high solution transferability~\cite{bardenet2013collaborative}, the weight estimation approach in~\cite{min2019multiproblem} is prone to underestimate their similarity and miss out on the potentially valuable knowledge accordingly.
Besides, the lack of task adaptation makes the multi-problem surrogates less effective in solving problems with strong heterogeneity.
To address this issue, Min \emph{et al.}~\cite{min2021generalizing} proposed a generalized transfer Bayesian optimization algorithm, in which neural networks are employed to bridge the gap between the source and target tasks.
A cross-task covariance term is used for utilizing the evaluated samples from a source task to achieve better prediction performance on the target task.
However, it is only applicable to problems with a single source task and still suffers from the issue of model-dependency.

\subsection{Motivations Behind This Work}
In light of the above, we aim to develop a general sequential transfer technique for improving SAS, which is supposed to have the following three favorable merits:

\begin{itemize}
	\item [1)] \emph{Portability:} The knowledge transfer method should be plug and play (i.e., surrogate model-independent), enabling it to be coherent with the state-of-the-art in SAS.
	\item [2)] \emph{Reliability:} It should be equipped with rank-based similarity instead of the objective value-based one, as the latter is prone to underestimate the promising knowledge for problems containing tasks with heterogeneous orders of magnitude but high ordinal correlations.
   \item [3)] \emph{Adaptivity:} It should be equipped with task adaptation, enabling it to solve problems with heterogeneous tasks.
\end{itemize}

Moreover, we show particular interest in the lower bound of the convergence gain brought by the proposed method and the important conditions leading to positive convergence gain, which will be theoretically derived in Section \ref{section:the}.


\section{Proposed Methods}
\label{section:ckt}

Fig. \ref{fig:flowchart} shows the flow chart of the proposed SAS-CKT, which aims to improve the optimization performance of SAS with a plug and play module named competitive knowledge transfer.
Unlike the traditional SAS that acquires promising solutions using the target surrogate model only, SAS-CKT has a knowledge competition module that treats both the source and target solutions as task-solving knowledge and compares them from a consistent view.

As shown in Fig. \ref{fig:flowchart}, the competitive knowledge transfer module consists of three key parts: \emph{similarity quantification}, \emph{task adaptation}, and \emph{knowledge competition}.
Specifically, similarity quantification is responsible for measuring the similarities between the source and target tasks while task adaptation is for improving the transferability of knowledge in case of task heterogeneity.
Lastly, knowledge competition makes the promising solutions acquired by the target surrogate compete with the optimized solutions of the source tasks, allowing one to identify the winner for true evaluation.
In what follows, the three key parts are introduced in detail.

\subsection{Similarity Quantification}
Suppose the database of a target EOP is denoted by $\mathcal{D}=\lbrack X,\boldsymbol{y}\rbrack$ and a knowledge base with $k$ previously-optimized EOPs is available, we denote the source surrogate models as $\hat{f}^s_i\left(\boldsymbol{x}\mid X_i,\boldsymbol{y}_i\right),\,1\le i\le k$.
Without loss of generality, we employ the Spearman's rank correlation to measure the similarity between the source and target tasks due to its efficient utilization of the solution data and widespread application in transfer optimization, whose superiority against other metrics will be empirically validated in Section \ref{subsection:reliability}.
Since the source surrogates normally have wider coverage over the search space\footnote{For source and target tasks with distinct decision spaces, one can transform them into a common search space $\Omega_c=\lbrack0,1\rbrack^d$ for knowledge transfer.} as compared to the target surrogate built upon limited data, their prediction performance is generally better than the target one.
Therefore, we propose surrogate-based Spearman's rank correlation (SSRC) to measure the source-target similarities based on the true responses of the evaluated solutions in $\mathcal{D}$ and their predicted responses on the source surrogates, as given by
\begin{equation}
s_i =\frac{\mathrm{cov}\left(\mathcal{R}\lbrack\hat{\boldsymbol{y}}_i\rbrack,\mathcal{R}\lbrack\boldsymbol{y}\rbrack\right)}{\mathrm{std}\left(\mathcal{R}\lbrack\hat{\boldsymbol{y}}_i\rbrack\right)\mathrm{std}\left(\mathcal{R}\lbrack\boldsymbol{y}\rbrack\right)},
\label{eq:similarity}
\end{equation}
where $s_i\in\lbrack-1,1\rbrack$ is the similarity between the $i$-th source task and the target task, $\mathcal{R}\lbrack\boldsymbol{a}\rbrack$ is the rank vector of $\boldsymbol{a}$, $\hat{\boldsymbol{y}}_i$ represents the predicted responses of $X$ on the $i$-th source surrogate.

\subsection{Task Adaptation}

For problems with heterogeneous tasks, task adaptation provides a solution for bridging the task gap and thus improving the transferability of knowledge.
Without loss of generality, we propose to adapt the source solutions using a transformation $\phi\left(\boldsymbol{x}\right):\Omega_s\to\Omega_t$, such that $\boldsymbol{x}'=\phi\left(\boldsymbol{x}\right)$.
Suppose $\phi$ is bijective and its inverse is $\phi^{-1}:\Omega_t\to\Omega_s$, the inversely adapted target solutions are denoted as $X^\circ=\phi^{-1}\left(X\right)$.
It is noted that improving the transferability depends on what is used for representing (or indicating) the transferability~\cite{xue2023solution}.
Following this clue, we propose surrogate-based domain adaptation (SDA) to improve the transferability reflected by Eq. \eqref{eq:similarity} with an optimal mapping, as formulated by

\begin{equation}
\begin{split}
\tilde{\phi}_i=\max_{\phi}s^a_i=\max_{\phi}\alpha_{\phi}\frac{\mathrm{cov}\left(\mathcal{R}\lbrack\hat{\boldsymbol{y}}^\circ_i\rbrack,\mathcal{R}\lbrack\boldsymbol{y}\rbrack\right)}{\mathrm{std}\left(\mathcal{R}\lbrack\hat{\boldsymbol{y}}^\circ_i\rbrack\right)\mathrm{std}\left(\mathcal{R}\lbrack\boldsymbol{y}\rbrack\right)},
\end{split}
\label{eq:adaptation}
\end{equation}
where $\tilde{\phi}_i$ denotes the optimal mapping for adapting the $i$-th source task, $s^a_i$ is the adaptation-based similarity, $\alpha_{\phi}$ is a regularization term that prevents the mapping $\phi$ from overfitting, $\hat{\boldsymbol{y}}^\circ_i$ denotes the predicted responses of $X^\circ$ on the $i$-th source surrogate. Then, the optimized solution $\tilde{\boldsymbol{x}}_i\in X_i$ is adapted and prepared for knowledge transfer, i.e., $\tilde{\boldsymbol{x}}'_i=\tilde{\phi}_i\left(\tilde{\boldsymbol{x}}_i\right)$.

Without loss of generality, we consider the translation transformation in this work.
Then, the mapping can be explicitly parameterized as $\phi_{\boldsymbol{\theta}}\left(\boldsymbol{x}\right)=\boldsymbol{x}+\boldsymbol{\theta}$, where $\boldsymbol{\theta}$ is a translation vector.
The parameterized mapping $\phi_{\boldsymbol{\theta}}$ is bijective and its inverse is $\phi_{\boldsymbol{\theta}}^{-1}\left(\boldsymbol{x}'\right)=\boldsymbol{x}'-\boldsymbol{\theta}$.
Therefore, we can formulate the adaptation model in Eq. \eqref{eq:adaptation} into the following form:
\begin{equation}
\begin{split}
\tilde{\boldsymbol{\theta}}_i=\max_{\boldsymbol{\theta}}&\alpha_{\boldsymbol{\theta}}\frac{\mathrm{cov}\left(\mathcal{R}\lbrack\hat{f}^s_i\left(X-\Theta\right)\rbrack,\mathcal{R}\lbrack \boldsymbol{y}\rbrack\right)}{\mathrm{std}\left(\mathcal{R}\lbrack\hat{f}^s_i\left(X-\Theta\right)\rbrack\right)\mathrm{std}\left(\mathcal{R}\lbrack\boldsymbol{y}\rbrack\right)},\\
&\mathrm{subject\,\,to}\,\,\,\boldsymbol{\theta}\in\lbrack-1,1\rbrack^d,
\end{split}
\label{eq:ada_translation}
\end{equation}
where $\tilde{\boldsymbol{\theta}}_i$ is the estimated translation vector for the $i$-th source task, $\Theta$ is the broadcast matrix of $\boldsymbol{\theta}$ whose shape is compatible with $X$.
Without loss of generality, the regularization term $\alpha_{\boldsymbol{\theta}}$ is set to be $1-\mid\mid\boldsymbol{\theta}\mid\mid_{\infty}$.
The optimized solution from the $i$-th source task can be adapted as $\tilde{\boldsymbol{x}}'_i=\tilde{\boldsymbol{x}}_i+\tilde{\boldsymbol{\theta}}_i$.

It should be noted that solving the problem in Eq. \eqref{eq:ada_translation}  requires some evaluations of the source surrogate models.
For most EOPs, such surrogate evaluations are typically computationally affordable, making it worth spending a certain number of surrogate evaluations solving the adaptation problem.
Moreover, it is worth noting that the task adaptation can be fulfilled in either an online or an offline fashion.
In the online manner, the task adaptation is supposed to be executed once the database of the target task is updated.
Alternatively, one can execute the task adaptation only once with the initial target data, which is computationally efficient but at the expense of less accurate adaptations.
This trade-off depends on the available computational resources allocated to the task adaptation.
Nevertheless, in Section \ref{section:the}, we show that the convergence gain brought by the CKT module without the task adaptation is bounded by zero regardless of source-target similarities.
In this sense, the task adaptation is more of an auxiliary component than an essential part of SAS-CKT, which is also empirically verified in Section \ref{subsection:adaptivity}.

\subsection{Competitive Knowledge Transfer (CKT)}
From the standpoint of SAS on the target task, we define internal and external improvements as follows:
\begin{itemize}
	\item \emph{Internal improvement:} the estimated (or predicted) improvement brought by a promising solution acquired by SAS on the target task under a particular infill criterion.
	\item \emph{External improvement:} the estimated (or predicted) improvement made by the optimized solution from a specific source task on the target task.
\end{itemize}

For a single run of the surrogate-assisted search, the interval improvement can be estimated as follows:
\begin{equation}
\Delta_{\mathrm{in}} = \min\lbrack\mathcal{I}\left(\mathcal{D}\right)\rbrack - \mathcal{I}\left(\boldsymbol{x}_p\right),
\label{eq:internal}
\end{equation}
where $\Delta_{\mathrm{in}}$ is the internal improvement, $\mathcal{I}\left(\cdot\right)$ represents the infill criterion (i.e., acquisition function) to be minimized on the target surrogate, $\boldsymbol{x}_p$ is the most promising solution obtained by SAS with a particular search algorithm on the target surrogate.
It is noted that the estimation in Eq. \eqref{eq:internal} is fairly general, which is compatible with many state-of-the-art SAS algorithms\footnote{Detailed calculations of the internal improvement under a variety of infill criteria are provided in Subsection A of the supplementary document.}, such as those listed in Table \ref{tab:sas_methods}.

For the external improvement, one may simply calculate it with Eq. \eqref{eq:internal} by replacing $\boldsymbol{x}_p$ with the optimized solution from a source task.
However, this strategy is problematic as the source optimized solution is very likely to be mistakenly estimated by the target surrogate trained on limited data, which is especially the case at the early stage of SAS.
In view of the fact that the source tasks had been optimized and their surrogates trained on a large number of evaluated solutions are relatively more reliable, we propose a convergence analogy-based estimation (CAE) method to estimate the external improvement.
Unlike the replacement strategy that relies on a potentially inaccurate target surrogate, CAE calculates improvements on the source surrogates and then analogizes them to the external improvements on the target task.
To enable the analogy, we postulate that the best objective values found by SAS on an optimization task are subject to the exponential decay due to its monotonicity and asymptotic convergence,
\begin{equation}
\gamma_\tau=\gamma_o+\gamma_i e^{-\lambda\tau},
\label{eq:exp_decay}
\end{equation}
where $\gamma_{\tau}$ is the best objective value ever found till time $\tau$, $\gamma_i$ and $\gamma_o$ represent the initial and optimal objective values, respectively, $\lambda$ denotes the decay constant.
For a specific task, its optimization trace can be modeled by Eq. \eqref{eq:exp_decay} with particular parameters estimated by the ordinary least squares method.

For the $i$-th source task, the improvement brought by its optimized solution against the target solutions is estimated as
\begin{equation}
\Delta^i_{y}=\min\left(\hat{\boldsymbol{y}}_i\right)-\min\left(\boldsymbol{y}_i\right),
\label{eq:improvement_y}
\end{equation}
where $\hat{\boldsymbol{y}}_i$ denotes the predicted responses of $X$ on the $i$-th source surrogate, $\boldsymbol{y}_i$ represents the objective values of all the evaluated solutions from the $i$-th source task.
It should be noted that the improvement in Eq. \eqref{eq:improvement_y} is in the context of the $i$-th source task and cannot be used as the improvement on the target task, as their orders of magnitude may vary a lot.
To address this issue, CAE employs the exponential decay in Eq. \eqref{eq:exp_decay} to analogize the improvement using the time interval corresponding to $\Delta^i_{y}$, which is given by
\begin{equation}
\begin{split}
\Delta^i_{\tau}&=\tau\lbrack\min\left(\boldsymbol{y}_i\right)\rbrack-\tau\lbrack\min\left(\hat{\boldsymbol{y}}_i\right)\rbrack\\
&=\tau^i_{\mathrm{max}}-\mathrm{ln}\left(\frac{\gamma^{i}_i}{\min\left(\hat{\boldsymbol{y}}_i\right)-\gamma_o^i}\right)/\lambda^i\\
&=\tau^i_{\mathrm{max}}-\tau^i_v,
\end{split}
\label{eq:improvement_tau}
\end{equation}
where $\tau^i_{\mathrm{max}}$ denotes the time required by SAS to achieve the objective value of $\min\left(\boldsymbol{y}_i\right)$ on the $i$-th source task, which is approximately equal to the maximum number of evaluations, $\gamma^{i}_i$ and $\gamma^{i}_o$ are the estimated initial and optimal objective values of the $i$-th source exponential decay, $\lambda^i$ is the estimated decay constant, $\tau^i_v$ is the estimated time required by SAS to achieve the objective value of $\min\left(\hat{\boldsymbol{y}}_i\right)$ on the $i$-th source task.

\begin{algorithm}
    \SetKwInOut{Input}{Input}
    \SetKwInOut{Output}{Output}
    \SetNoFillComment
    \caption{$\mathtt{SAS}$-$\mathtt{CKT}$}
    \label{alg:gl}
    \begin{small}
    \Input{$f\left(\boldsymbol{x}\right)$ (target task), $\mathcal{M}=\{\lbrack X_i,\boldsymbol{y}_i\rbrack,\,i=1,...,k\}$ (knowledge base), $\hat{f}^s_{1\le i\le k}\left(\boldsymbol{x}\right)$ (source surrogates)}
    \Output{$\boldsymbol{x}_{b}$ (the best solution)}
	\begin{tikzpicture}[remember picture,overlay]
    \draw [fill=myboxcolor, ultra thin] ($(pic cs:a) + (-0.93,0.27)$) rectangle ($(pic cs:b)+(7.9,0.34)$);
	\end{tikzpicture}
	$\mathcal{D}\leftarrow$Initialize a database using a random sampling method\;
	\While{termination condition is not met}
	{
		\tcc{Internal Improvement}
		$\lbrack\boldsymbol{x}_p,\Delta_{\mathrm{in}}\rbrack\leftarrow\mathtt{SAS}(\mathcal{D})$\;
		\tikzmark{a}\uIf{$\mathtt{CKT}$ is allowed, i.e., $\mathtt{mod}\left(|\mathcal{D}|,\delta\right)=0$}
		{
			\tcc{External Improvements}
			$\lbrack\gamma^t,\gamma^{1\le i\le k}\rbrack\leftarrow$ Fit the exponential decays\;
				\For{$i=1\to k$}
				{
					$\Delta^i_{\mathrm{ex}}\leftarrow$The external improvement in Eq. \eqref{eq:external}\;
				}
			\tcc{Knowledge Competition}
			\uIf{$\Delta_{\mathrm{in}}<\max_{1\le i\le k}\Delta^i_{\mathrm{ex}}$}
			{
				$\boldsymbol{x}_e\leftarrow\tilde{\boldsymbol{x}}_{\mathrm{max}}$\;
			}
			\Else
			{
				$\boldsymbol{x}_e\leftarrow\boldsymbol{x}_p$\;
			}
		}\tikzmark{b}
		\Else
		{
			$\boldsymbol{x}_e\leftarrow\boldsymbol{x}_p$\;
		}
		\tcc{Update of the Database}
		$\mathcal{D}\leftarrow\mathcal{D}\cup\lbrack\boldsymbol{x}_e,f\left(\boldsymbol{x}_e\right)\rbrack$\;
	}
	\textbf{return} the best solution $\boldsymbol{x}_b\in\mathcal{D}$\;
    \end{small}
\end{algorithm}

When analogizing the improvement on the target task with the improvement on the $i$-th source task, their similarity should be taken into account.
When $s_i$ is 0, the source optimized solution (i.e., $\tilde{\boldsymbol{x}}_i$) can be seen as a randomly generated solution for the target task.
In this case, the expected time required by SAS to achieve the quality of $\tilde{\boldsymbol{x}}_i$ is simply 0.
When $s_i$ is 1, it can be concluded that the two tasks are very similar, enabling one to use the improvement on the source task to infer the external improvement brought by $\tilde{\boldsymbol{x}}_i$.
In summary, the confidence in the analogized time of $\tilde{\boldsymbol{x}}_i$ on the target task (i.e., $\tau+\Delta^i_\tau$) grows with the similarity $s_i$, where $\tau$ is the current time of the target search, which is simply the number of used function evaluations.
Without loss of generality, CAE uses the linear correlation between $s_i$ and $\tau+\Delta^i_\tau$ to estimate the external improvement brought by $\tilde{\boldsymbol{x}}_i$, which is given by

\begin{equation}
\Delta^i_{\mathrm{ex}} = s_i^+\lbrack\min\left(\boldsymbol{y}\right)-\gamma^t_{s_i\left(\tau+\Delta^i_\tau\right)}\rbrack,
\label{eq:external}
\end{equation}
where $\Delta^i_{\mathrm{ex}}$ denotes the external improvement brought by the optimized solution from the $i$-th source task (i.e., $\tilde{\boldsymbol{x}}_i$), $s_i^+$ is the rectified value of $s_i$, i.e., $s_i^+=\max\left(0,s_i\right)$, which is used for suppressing the knowledge transfer when $s_i<0$, $\gamma^t\left(\cdot\right)$ denotes the exponential decay model of the target task.

Finally, the knowledge competition kicks in by comparing the internal and external improvements, enabling the winner to undergo the real evaluation, which is given by

\begin{equation}
\boldsymbol{x}_e=
\begin{cases}
\tilde{\boldsymbol{x}}_{\mathrm{max}},&\Delta_{\mathrm{in}}<\Delta^{\mathrm{max}}_{\mathrm{ex}},\\
\boldsymbol{x}_p,&\Delta_{\mathrm{in}}\ge\Delta^{\mathrm{max}}_{\mathrm{ex}},
\end{cases}
\label{eq:promising_solution}
\end{equation}
where $\boldsymbol{x}_e$ is the candidate solution to be evaluated, $\Delta^{\mathrm{max}}_{\mathrm{ex}}=\max_{1\le i\le k}\Delta^i_{\mathrm{ex}}$ denotes the maximum external improvement, $\tilde{\boldsymbol{x}}_{\mathrm{max}}$ is the optimized solution of the source task with the maximum external improvement.
When using the task adaptation, one only needs to replace the similarity in Eq. \eqref{eq:external} by the adaptation-based similarity in Eq. \eqref{eq:adaptation} and adapt the source solutions, with the other procedures remained unaltered.

\subsection{Implementation of SAS-CKT}

The pseudo code of SAS-CKT is presented in Algorithm 1.
The framed knowledge competition phase shown in lines 4 to 11 distinguishes SAS-CKT from the traditional SAS.
The good portability of the proposed method enables it to boost the performance of various SAS engines.
For an SAS-CKT algorithm equipped with a specific SAS engine, the knowledge competition can be uninterruptedly triggered by comparing the internal and external improvements at each real evaluation.
However, the landscape update of the target task is usually mild when only one newly evaluated solution is archived, as shown in line 14, making the external improvements bear insignificant changes.
Therefore, a fixed internal $\delta$ in terms of real evaluation for triggering the knowledge competition can be used to reduce its frequent calculations, as shown in line 4.
Lastly, it should be noted that the use of the task adaptation will not change the implementation of SAS-CKT.
One only needs to replace the similarity by the adaptation-based similarity when estimating the external improvements in line 7 and adapt the source optimized solution with the learned mapping for knowledge transfer in line 9.

\subsection{Computational Complexity}

According to Algorithm \ref{alg:gl}, we find that the computational complexity of the proposed CKT method mainly comes from the estimation of the external improvements.
This process involves calling the $k$ source surrogate models for similarity quantification with Eq. \eqref{eq:similarity}, whose overall complexity is $O\left(kn\cdot c_{\mathrm{p}}\right)$, where $n$ represents the number of evaluated target solutions, $c_{\mathrm{p}}$ is the prediction cost of source surrogate.
With respect to both $k$ and $n$, CKT exhibits linear computational complexity.
As for the prediction cost $c_{\mathrm{p}}$ in SAS-CKT, it depends on the source surrogate used.
For most surrogate models, their prediction costs are far lower than the training costs.
For instance, the prediction complexities of commonly used surrogates, such as RBF and GPR used in our experiments, are simply $O\left(dN\right)$, where $d$ denotes the problem dimension, $N$ is the number of training samples of the source surrogates.


\section{Theoretical Analyses}
\label{section:the}

In this section, we conduct theoretical analyses of SAS-CKT.
Firstly, two prerequisites and three supporting lemmas are briefly discussed.
Then, the lower bound of the expected convergence gain brought by the knowledge competition is mathematically analyzed.
Lastly, a brief discussion about the implications of the derived theoretical results is presented.

\subsection{Supporting Lemmas}

There are two prerequisites in SAS-CKT: 1) the credibility of optimization experience hidden in the source tasks and 2) the sufficiency of the similarity for the optimum equivalence:
\begin{itemize}
\item The credibility of optimization experience ensures the improvement in Eq. \eqref{eq:improvement_y} for analogizing the external improvement, which can be solidified further if the global convergence of SAS is proved.
It also fulfills the optimality needed by the optimum equivalence.

\item The sufficiency of the similarity for the optimum equivalence is essential for the analogy: the entire improvement on the source task is used tor infer the external improvement if the similarity value is one, as the optima of the source and target tasks are deemed identical in this case.
\end{itemize}

\begin{figure}[ht]
	\centering
	\includegraphics[width=2.5in]{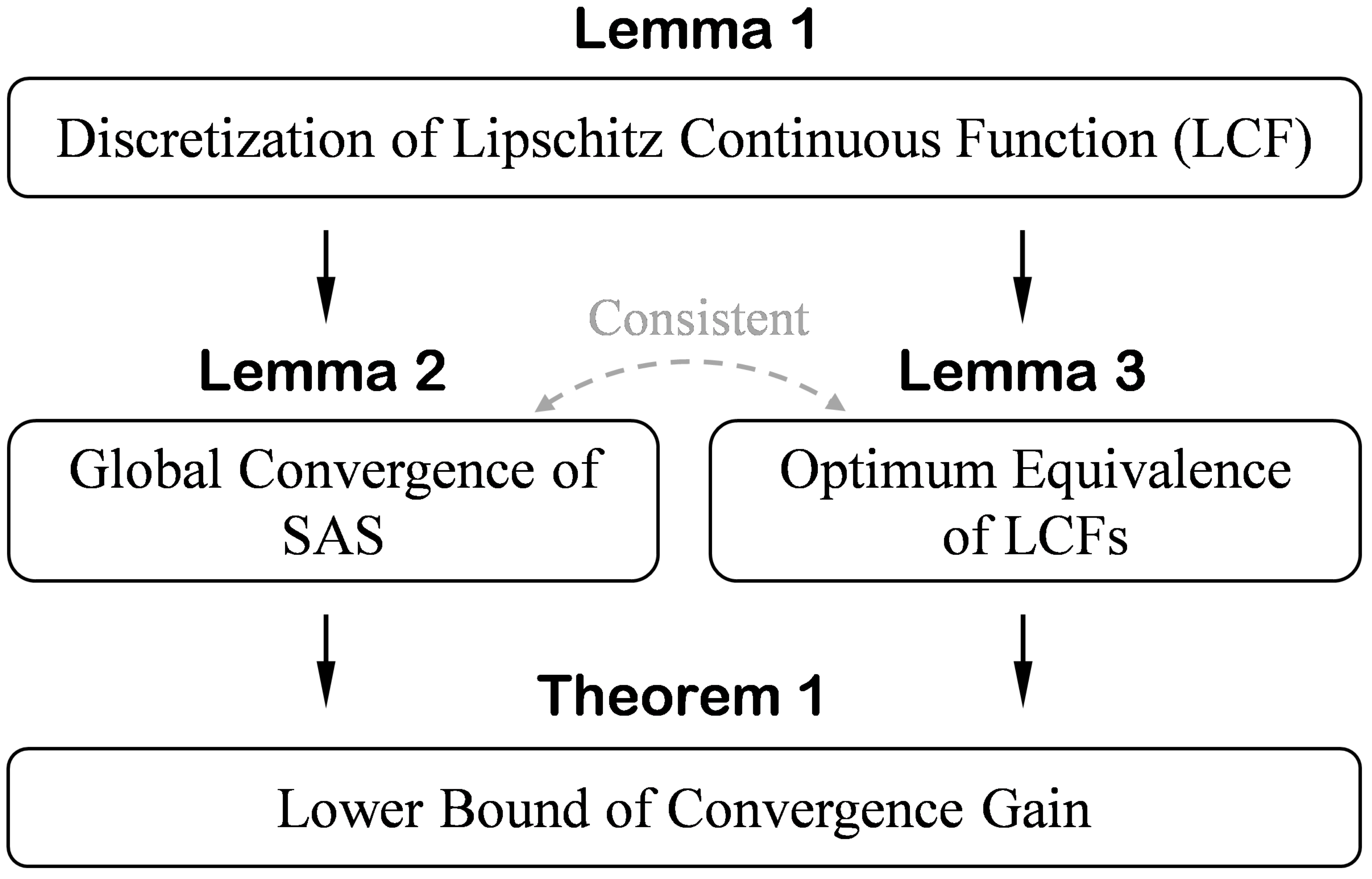}
	\caption{Relations between the three lemmas and Theorem 1.}
	\label{fig:theorems}
\end{figure}

With the Lipschitz condition, the satisfaction of the two prerequisites can be analytically proved.
Firstly, the discretization of a Lipschitz continuous function is presented in Lemma 1, enabling us to deal with it in a finite search space.
Then, the global convergence of SAS for the credibility of optimization experience and the sufficiency of the similarity for the optimum equivalence are confirmed by Lemma 2 and Lemma 3, respectively.
Lastly, the lower bound of the expected convergence gain brought by the knowledge competition is mathematically derived, with the analysis of positive convergence gain presented in Theorem 1.
Fig. \ref{fig:theorems} sketches the relations between the three lemmas and the theorem.
Due to the page limit, the three lemmas are provided in Subsection B of the supplementary document accompanying this paper.

\subsection{Analysis of Convergence Gain}

It is assumed that the best objective values searched by SAS are subject to the exponential decay\footnote{The empirical evidence for supporting this postulation and the parameter estimation are provided in Subsection C of the supplement.}, which is given by
\begin{equation}
\gamma_{\tau}=\gamma_o + \gamma_i e^{-\lambda\tau}.
\notag
\end{equation}

Suppose the exponential decay of the target task is given by $\gamma^t\left(\tau;\gamma^t_o,\gamma^t_i,\lambda^t\right)$, we can estimate the expected internal improvement from $\tau$ to $\tau+1$ as follows:
\begin{equation}
\Delta_{\mathrm{in}} = \gamma^t_{\tau}-\gamma^t_{\tau+1}.
\notag
\end{equation}

For the $i$-th source task characterized by $\gamma^i\left(\tau;\gamma^i_o,\gamma^i_i,\lambda^i\right)$, the expected external improvement brought by the optimized solution can be estimated as follows:
\begin{equation}
\Delta^i_{\mathrm{ex}} = s_i^+\left(\gamma^t_{\tau}-\gamma^t_{s_i\left(\tau+\Delta^i_\tau\right)}\right).
\notag
\end{equation}

\begin{figure*}[ht]
	\centering
	\includegraphics[width=6.9in]{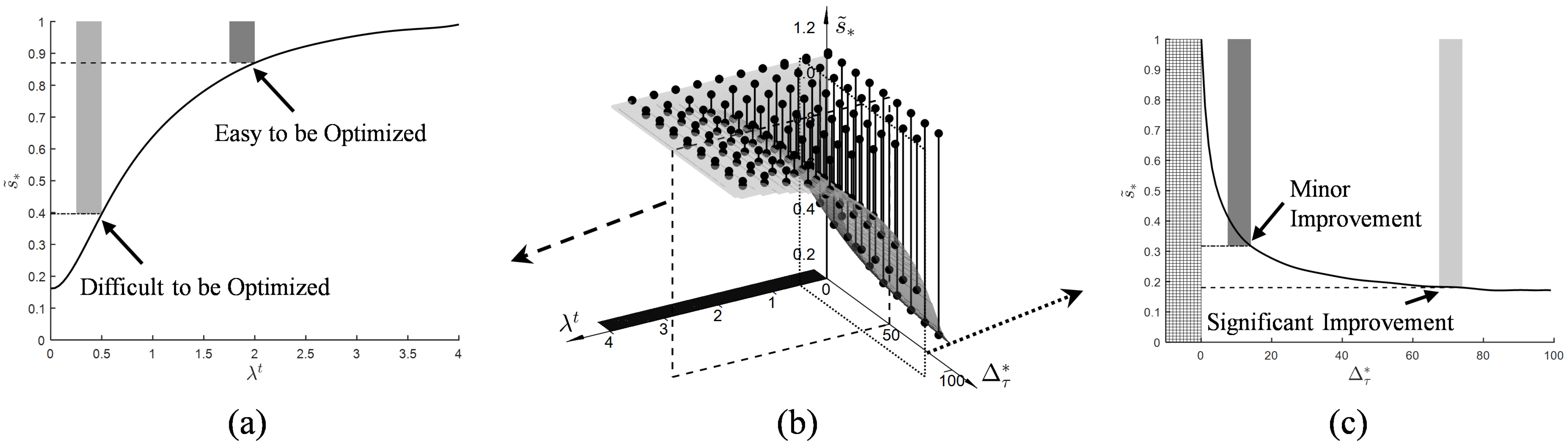}
	\caption{Illustration of the impacts of $\lambda^t$ and $\Delta^*_\tau$ on the lower bound of the conditional similarity for positive convergence gain (i.e., $\tilde{s}_*$): (a) the lower bound of the similarity with respect to $\lambda^t$ when $\Delta^*_\tau=50$; (b) a 3D surface for demonstrating the joint impact of $\lambda^t$ and $\Delta^*_\tau$ on $\tilde{s}_*$; (c) the lower bound of the similarity with respect to $\Delta^*_\tau$ when $\lambda^t=0.16$.}
	\label{fig:positive_transfer}
\end{figure*}

Given a knowledge base with $k$ source tasks, the convergence gain brought by CKT can be estimated as follows:
\begin{equation}
\begin{split}
\mathcal{C}\left(s_*\right) &= \max\lbrace\Delta_{\mathrm{in}},\Delta^{\mathrm{max}}_{\mathrm{ex}}\rbrace-\Delta_{\mathrm{in}}\\
&=\max\lbrace0,s^+_*\cdot\lbrack\gamma^t_{\tau}-\gamma^t_{s_*\left(\tau+\Delta^*_\tau\right)}\rbrack-\lbrack\gamma^t_{\tau}-\gamma^t_{\tau+1}\rbrack\rbrace\\
&=\max\lbrace0,\Psi\left(s_*,\gamma^t_o,\gamma^t_i,\lambda^t,\tau,\Delta_\tau^*\right)\rbrace,
\label{eq:convergence_gain}
\end{split}
\end{equation}
where $s_*$ is the similarity of the source task with the maximum external improvement, which is termed \emph{conditional similarity} for short throughout this paper as it is conditioned on the improvement, $\Delta_\tau^*$ denotes the estimated time interval.
Suppose $s_*$ of a knowledge base is a random variable, we define the expected convergence gain brought by CKT as follows:
\begin{defi} For a target task and a knowledge base with the conditional similarity under a probability density function $f\left(s_*\right)$, the expected convergence gain brought by the knowledge competition to the target task at time $\tau$ is defined as
\begin{equation}
\mathbb{E}\lbrack\mathcal{C}\left(s_*\right)\rbrack = \int_{-1}^1\mathcal{C}\left(s_*\right)f\left(s_*\right)\mathrm{d}s_*= \int_{0}^1\mathcal{C}^+\left(s_*\right)f\left(s_*\right)\mathrm{d}s_*,
\notag
\end{equation}
where $\mathcal{C}^+=\max\lbrace0,\Psi\left(s_*,\gamma^t_o,\gamma^t_i,\lambda^t,\tau,\Delta^*_\tau\right)\rbrace$ with $s_*\ge0$.
\end{defi}

It can be seen that the lower bound of the convergence gain is bounded by zero regardless of $f\left(s_*\right)$.
Moreover, we show particular interest in conditions leading to positive convergence gain.
A theorem in this respect is presented as follows:

\begin{thm}
Given a target task and a knowledge base, the lower bound of the expected convergence gain brought by the knowledge competition is greater than zero if $f\left(s_*\right)$ of the knowledge base is bounded from zero within $\lbrack\tilde{s}_*,1\rbrack$, where $\tilde{s}_*$ is the root of the following equation:
\begin{equation}
\Psi\left(\tilde{s}_*,\gamma^t_o,\gamma^t_i,\lambda^t,\tau,\Delta^*_\tau\right)=0.
\notag
\end{equation}
\end{thm}

\begin{IEEEproof}[Proof] With the non-negativity of $\mathcal{C}^+$, we have
\begin{equation}
\mathbb{E}\lbrack\mathcal{C}\left(s_*\right)\rbrack=\sum_{i=1}^\nu\int_{\underline{s}_*^i}^{\overline{s}_*^i} \Psi\left(s_*,\gamma^t_o,\gamma^t_i,\lambda^t,\tau,\Delta^*_\tau\right)f\left(s_*\right)\mathrm{d}s_*,
\notag
\end{equation}
where $\nu$ is the total number of intervals in which $\Psi$ is greater than 0, $\underline{s}_*^i$ and $\overline{s}_*^i$ denote the lower and upper bounds of the $i$-th interval, respectively.

Given the known parameters $\boldsymbol{\zeta}=\lbrace\gamma^t_o,\gamma^t_i,\lambda^t,\tau,\Delta^*_\tau\rbrace$, we denote $\Psi$ with respect to $s_*$ as $\psi\left(s_*;\boldsymbol{\zeta}\right)$.
Then, let us examine the monotonicity of $\psi$ according to its first-order derivative:

\begin{equation}
\frac{\mathrm{d}\psi}{\mathrm{d}s_*}=s_*\lambda^t\left(\tau+\Delta^*_\tau\right)\gamma^t_{s_*\left(\tau+\Delta^*_\tau\right)}+\lbrack\gamma^t_{\tau}-\gamma^t_{s_*\left(\tau+\Delta^*_\tau\right)}\rbrack.
\label{eq:first_derivative}
\end{equation}

According to Eq. \eqref{eq:convergence_gain}, we find that the non-negativity of the convergence gain depends on the non-negativity of $s_*$ and $\Delta^*_\tau$.
Then, with Eq. \eqref{eq:first_derivative}, we can deduce that the first derivative of $\psi$ is non-negative if the following condition holds:
\begin{equation}
\tau<s_*\left(\tau+\Delta^*_\tau\right) \Rightarrow s_*>\frac{\tau}{\tau+\Delta^*_\tau}.
\notag
\end{equation}

With the global convergence of the source tasks, we have
\begin{equation}
\lim_{\tau_{\mathrm{max}}\to\infty} \frac{\tau}{\tau+\Delta^*_\tau}=\lim_{\tau_{\mathrm{max}}\to\infty} \frac{\tau}{\tau+\tau_{\mathrm{max}}-\tau^*_v}=0.
\notag
\end{equation}

The non-negativity of the first-order derivative within $(0,1]$ indicates that $\psi$ is strictly increasing on $(0,1]$.
Thus, we have 
\begin{equation}
\frac{\mathrm{d}\psi}{\mathrm{d}s_*}>0,\,\forall s_*\in(0,1] \Rightarrow \psi\left(s_*;\boldsymbol{\zeta}\right)>0,\,\forall s_*\in(\tilde{s}_*,1].
\notag
\end{equation}

Then, we can reformulate the expected convergence gain as follows:
\begin{equation}
\begin{split}
\mathbb{E}\lbrack\mathcal{C}\left(s_*\right)\rbrack &=\int_{0}^{\tilde{s}_*}\mathcal{C}^+\left(s_*\right)f\left(s_*\right)\mathrm{d}s_*+\int_{\tilde{s}_*}^{1}\mathcal{C}^+\left(s_*\right)f\left(s_*\right)\mathrm{d}s_*\\
&=\int_{\tilde{s}_*}^{1}\psi\left(s_*;\boldsymbol{\zeta}\right)f\left(s_*\right)\mathrm{d}s_*.
\end{split}
\notag
\end{equation}

Since $\psi\left(s_*;\boldsymbol{\zeta}\right)$ is non-negative and $f\left(s_*\right)$ is bounded from zero within $\lbrack\tilde{s}_*,1\rbrack$, we have
\begin{equation}
\mathbb{E}\lbrack\mathcal{C}\left(s_*\right)\rbrack =\int_{\tilde{s}_*}^{1}\psi\left(s_*;\boldsymbol{\zeta}\right)f\left(s_*\right)\mathrm{d}s_*>0.
\notag
\end{equation}

\end{IEEEproof}

\subsection{A Brief Discussion}
In this part, we will briefly discuss the implications of Theorem 1.
Firstly, it should be noted that $\gamma^t_o$ and $\gamma^t_i$ do not influence the lower bound of the similarity for positive convergence gain.
When analyzing the impacts of $\lambda^t$, $\tau$ and $\Delta^*_\tau$ on $\tilde{s}_*$ using the implicit term $\Psi$, we set $\tau$ to 1 for the sake of visualizing the joint impact of $\lambda^t$ and $\Delta^*_\tau$.
The impact of $\tau$ will be separately analyzed later.
Fig. \ref{fig:positive_transfer} illustrates the impacts of $\lambda^t$ and $\Delta^*_\tau$ on the lower bound of the conditional similarity for positive convergence gain, as presented in Fig. \ref{fig:positive_transfer}(a) and Fig. \ref{fig:positive_transfer}(c), respectively.
The joint impact is demonstrated by a 3D surface shown in Fig. \ref{fig:positive_transfer}(b), in which the lines with endpoints are used for indicating the similarity intervals for positive convergence gain (i.e., $\lbrack\tilde{s}_*,1\rbrack$) and the black band for $\Delta^*_\tau<1$ signifies the cancellation of knowledge transfer.

To separately examine the impacts of $\lambda^t$ and $\Delta^*_\tau$, we investigate each of the two parameters by fixing the other one.
Specifically, $\lambda^t$ and $\Delta^*_\tau$ are fixed to 0.16 and 50, respectively, as indicated by the dotted frames in Fig. \ref{fig:positive_transfer}(b).
For the impact of $\lambda^t$ shown in Fig. \ref{fig:positive_transfer}(a), we can observe the monotonically increasing trend of $\tilde{s}_*$ with respect to $\lambda^t$.
A question arises: what does the parameter $\lambda^t$ represent?
According to Eq. \eqref{eq:exp_decay}, we find that $\lambda$ can reflect the decay rate of an exponential decay model, i.e., the bigger $\lambda$ is, the faster the quantity vanishes.
Since we model the best objective values found by SAS as the exponential decay, $\lambda^t$ can reflect how difficult it is to optimize the target task, i.e., the bigger $\lambda^t$ is, the easier the target task is to be optimized.
Thus, the increasing trend in Fig. \ref{fig:positive_transfer}(a) makes intuitive sense: an easy task requires higher quality of knowledge (i.e., higher similarity) for positive convergence gain than a difficult task.
By contrast, we can observe the monotonically decreasing trend of $\tilde{s}_*$with respect to $\Delta^*_\tau$ in Fig. \ref{fig:positive_transfer}(c).
The parameter $\Delta^*_\tau$ indicates the potential improvement brought by the source optimized solution for the target search.
This observation makes intuitive sense: the less significant the potential improvement of knowledge is, the higher $\tilde{s}_*$ should be to achieve positive convergence gain.
The impact of $\tau$ on $\tilde{s}_*$ can be analyzed in the same way with the fixed $\lambda^t$ and $\Delta^*_\tau$.
The monotonically increasing trend of $\tilde{s}_*$ with respect to $\tau$ indicates that the quality of knowledge should grow as the target search proceeds.
This also makes intuitive sense: a solution that can speed up the target search at the early stage could be less promising or even harmful to the target search at the later stage, since the quality of target solutions gradually improves as the optimization proceeds.

Lastly, let us summarize the four important conditions for positive convergence gain, among which the first two conditions are target task-related while the last two conditions are knowledge base-related, as itemized as follows:
\begin{itemize}
\item \emph{Condition 1:} $\underline{\lambda^t<\infty}$. This requires that the target task will not reach its global optimum immediately, leaving space for knowledge transfer to speed up the target search.
A small $\lambda^t$ caused by either difficult problems or weak optimizers will reduce $\tilde{s}_*$ and make the knowledge transfer yield positive convergence gain more easily\footnote{The empirical evidence can be found in Section VII of~\cite{xue2023solution}.}.

\item \emph{Condition 2:} $\underline{\tau<\infty}$. This condition shows that the optimum of the target task has not been reached till time $\tau$, indicating that there is still room for improvement through knowledge transfer.
At the early optimization stage indicated by a small $\tau$, the knowledge transfer can produce positive convergence gain more easily.

\item \emph{Condition 3:} $\underline{\Delta^*_\tau>1}$. This requires that the potential improvement brought by the knowledge should be greater than the internal improvement brought by SAS, ensuring that the knowledge is qualified as the search experience from the source tasks to speed up the target search.

\begin{table}[ht]
	\caption{Wins, ties and losses of SAS-CKT against SAS with the 6 backbone optimizers on the 12 test problems.}
	\centering
	\footnotesize
	\heavyrulewidth=0.12em
	\lightrulewidth=0.1em
	\cmidrulewidth=0.1em
	\setlength\tabcolsep{3pt}
	\begin{tabular}{lcccc}
		\toprule
Optimizer&HS Problems&MS Problems&LS Problems&Summary\\
		\midrule
		BO-LCB~\cite{shahriari2015taking}&2/2/0&4/0/0&0/4/0&6/6/0\\
		\midrule
		TLRBF~\cite{li2021three}&3/1/0&4/0/0&1/3/0&8/4/0\\
		\midrule
		GL-SADE~\cite{wang2022surrogate}&0/4/0&1/3/0&0/4/0&1/11/0\\
		\midrule
		DDEA-MESS~\cite{yu2022data}&2/2/0&2/2/0&0/4/0&4/8/0\\
		\midrule
		LSADE~\cite{kuudela2023combining}&3/1/0&3/1/0&1/3/0&7/5/0\\
		\midrule
		AutoSAEA~\cite{xie2023surrogate}&3/1/0&4/0/0&0/4/0&7/5/0\\
		\midrule
		w/t/l (\%)&54/46/0&75/25/0&8/92/0&46/54/0\\
		\bottomrule
	\end{tabular}
	\label{tab:portability_results}
\end{table}

\item \emph{Condition 4:} $\underline{s_*\ge\tilde{s}_*}$. This condition requests the quality of knowledge to be transferred, enabling one to analogize the potential improvements on the source tasks to the target task for positive convergence gain.
\end{itemize}


\section{Experimental Setup}
\label{section:setup}

This section presents the setup of our experiments, including the test problems, the backbone SAS solvers, and the general parameter settings.

\subsection{Test Problems}
In this study, the proposed methods are validated on a benchmark suite in~\cite{xue2023scalable}, which contains 12 sequential transfer optimization problems (STOPs) with a broad spectrum of representation of the diverse similarity relationships between the optimal solutions of the source and target tasks in real-world problems.
This broad representation enables a more comprehensive evaluation of various knowledge transfer methods~\cite{xue2023solution,scott2023first}.
According to the similarity relationship, the 12 STOPs are divided into three categories: 1) high similarity (HS); 2) mixed similarity (MS); and 3) low similarity (LS).
In this study, the number of source tasks is set to 50.

\subsection{Backbone SAS Optimizers}
In this study, we employ six backbone optimizers to validate the portability of CKT, including a standard Bayesian optimizer~\cite{shahriari2015taking} (BO-LCB) and five state-of-the-art SAEAs listed in Table \ref{tab:sas_methods}, i.e., TLRBF~\cite{li2021three}, GL-SADE~\cite{wang2022surrogate}, DDEA-MESS~\cite{yu2022data}, LSADE~\cite{kuudela2023combining}, and AutoSAEA~\cite{xie2023surrogate}.
An SAS optimizer suffixed with CKT (e.g., AutoSAEA-CKT) represents its knowledge competition-based version without the task adaptation, while a backbone solver suffixed with CKT-A denotes its use of the adaptation-based knowledge competition.
For each STOP, its knowledge base is constructed by archiving the evaluated solutions obtained by BO-LCB on all the source tasks, i.e., $\mathcal{M}=\{\mathcal{D}^s_i: \lbrack X_i,\boldsymbol{y}_i\rbrack,\,i=1,...,k\}$.

\begin{figure}[ht]
	\centering
	\includegraphics[width=3.4in]{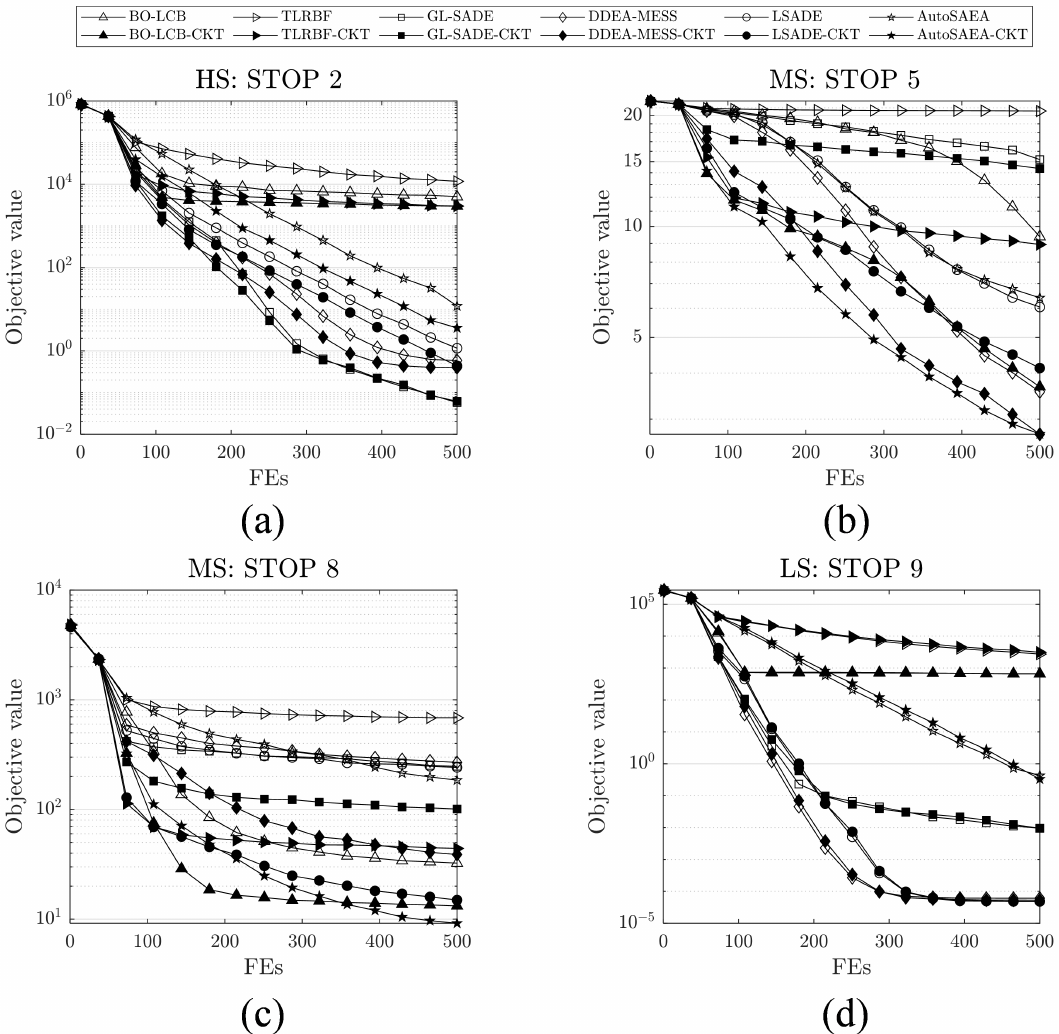}
	\caption{Averaged convergence curves of SAS-CKT against SAS with the six backbone optimizers on four problems: (a) STOP 2; (b) STOP 5; (c) STOP 8; (d) STOP 9.}
	\label{fig:portability_concurves}
\end{figure}

\subsection{Parameter Settings} The general parameter settings are listed as follows:
\begin{itemize}
	\item The number of initial solutions provided by LHS: 50.
	\item The population size of evolutionary solvers: 50.
	\item The number of iterations of the evolutionary search: 50.
	\item The maximum number of true function evaluations: 500.
	\item The transfer interval\footnote{The sensitivity analysis on this parameter is provided in the supplement.} in terms of real evaluation ($\delta$): 20.
	\item The number of independent runs: 30.
\end{itemize}

More detailed parameter settings of the backbone SAS optimizers are provided in Subsection D of the supplement.


\section{Experimental Results and Analyses}

\label{section:exp}

\subsection{Portability of CKT}

To demonstrate the portability of the proposed knowledge competition method, we summarize the wins, ties and losses of SAS-CKT against SAS with the six backbone optimizers based on the significance test of the best objective values in Table \ref{tab:portability_results}, whose detailed optimization results are provided in Subsection F of the supplementary document due to the page limit.
It can be seen that the knowledge competition yields a wide range of positive performance gain to the six backbone optimizers across the HS and MS problems while maintaining non-negative gain on all the LS problems, which validates its transfer learning capability to exhibit non-negative convergence gain when boosting different SAS engines.
This indicates that the knowledge competition is able to boost the performance of contemporary SAS optimizers automatically in cases of HS or MS problems while exhibiting reasonably comparable performance to the backbone optimizers in the worst-case scenario of LS problems.
Fig. \ref{fig:portability_concurves} shows the averaged convergence curves of the six backbone solvers and their knowledge competition-based counterparts on four selected problems.
It can be observed that the knowledge competition can achieve effective convergence speedups for the six SAS optimizes on the HS and MS problems while maintaining comparable convergence performance to the baseline solvers on the LS problems.
\begin{figure}[ht]
	\centering
	\includegraphics[width=3.4in]{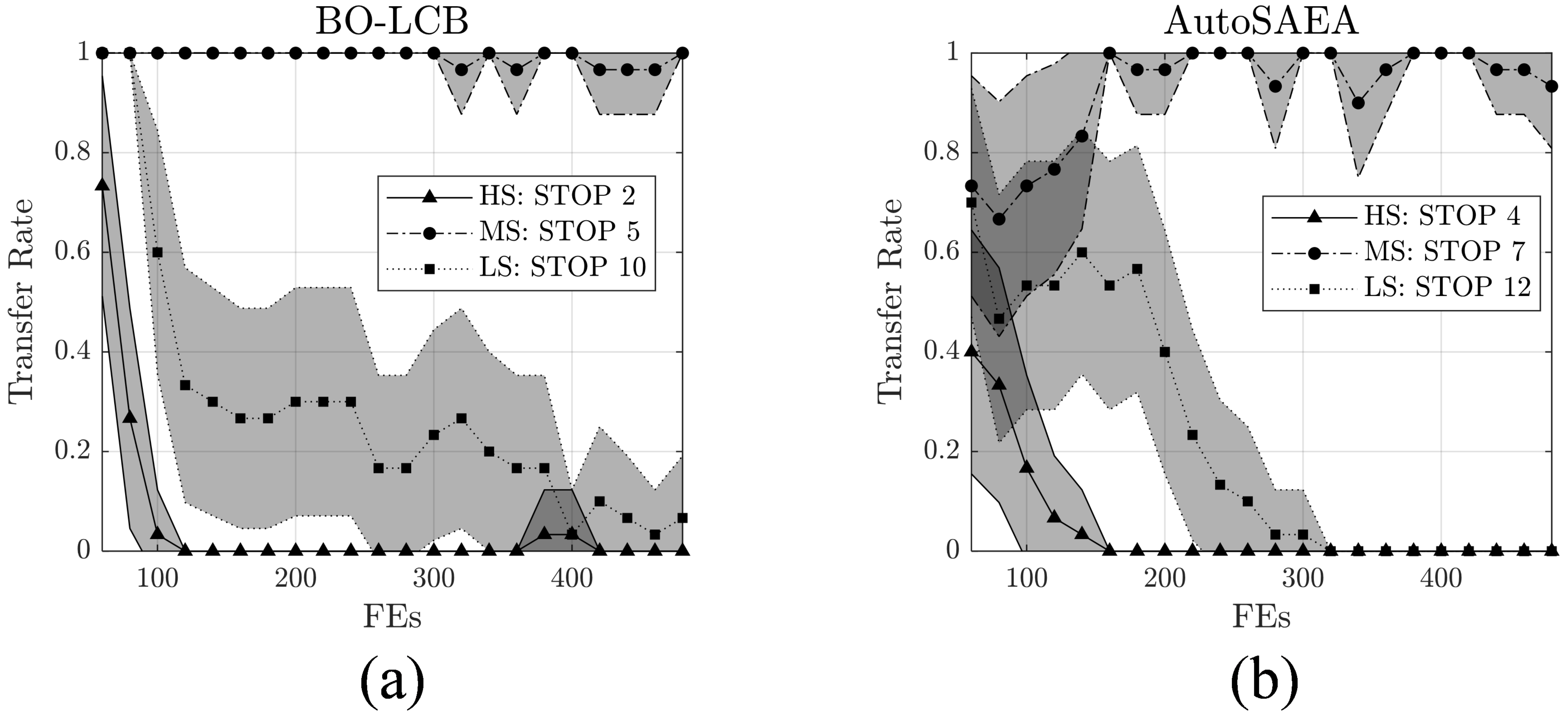}
	\caption{Averaged transfer rates of BO-LCB and AutoSAEA based on the knowledge competition on a few representative problems: (a) BO-LCB; (b) AutoSAEA. The shaded area spans 1/2 standard deviation on either side of the mean.}
	\label{fig:transfer_rates}
\end{figure}

To investigate the underlying transfer mechanism of SAS-CKT, we show the averaged transfer rates of BO-LCB-CKT and AutoSAEA-CKT over 30 independent runs on a number of representative problems in Fig. \ref{fig:transfer_rates}.
The transfer rate equals to one if $\Delta_{\mathrm{in}}<\Delta^{\mathrm{max}}_{\mathrm{ex}}$ while being zero otherwise.
It can be seen that the knowledge competition exhibits different patterns of transfer rates on the three types of problems.
Specifically, on the HS problems, we can observe the high transfer rates at the early stage, which are followed by their sharp declines.
The early high transfer rates are largely due to the identification of the highly similar source tasks, while the sharp declines can be attributed to the automatic closure of knowledge transfer by Eq. \eqref{eq:promising_solution} for avoiding the evaluation of similar optimized solutions with minor external improvements.
On the MS problems, we can observe the stably high transfer rates along the evolutionary search, which can be attributed to the variety of the optimized solutions from the source tasks with the mixed similarities to the target task.
As for the LS problems, the decreasing rates result from the adaptive suppression of transferring the optimized solutions from the source tasks with the low similarities to the target task.
Such adaptive closure of knowledge transfer makes SAS-CKT degrade into its backbone SAS, which explains their comparable performance on the LS problems.
In summary, the transfer mechanism of SAS-CKT fits the nature of STOPs with respect to the similarity relationship.

\subsection{Reliability of CKT}
\label{subsection:reliability}

To validate the reliability of CKT in similarity quantification, we first examine the superiority of the proposed surrogate-based Spearman's rank correlation (SSRC) over a number of similarity metrics commonly used in transfer optimization.
Then, SAS-CKT is compared with a state-of-the-art transfer algorithm for EOPs.

\subsubsection{Comparison with Other Similarity Metrics} We employ four similarity metrics, including the maximum mean discrepancy (MMD)~\cite{jiang2017transfer}, Wasserstein distance (WD)~\cite{zhang2019multisource}, mixture model (MM)~\cite{bali2019multifactorial}, and representation-based distance (RD)~\cite{yao2021self}, to compare with the proposed SSRC.
\begin{table}[ht]
	\caption{Wins, ties and losses of SSRC against the four similarity metrics under different FEs on the MS problems.}
	\centering
	\footnotesize
	\heavyrulewidth=0.12em
	\lightrulewidth=0.1em
	\cmidrulewidth=0.1em
	\setlength\tabcolsep{5.4pt}
	\begin{tabular}{lccccc}
		\toprule
\multirow{3}*{Metric}&\multicolumn{5}{c}{MS Problems}\\
		\cmidrule(){2-6}
		&FEs=100&FEs=200&FEs=300&FEs=400&FEs=500\\
		\midrule
		MMD~\cite{jiang2017transfer}&1/3/0&2/2/0&1/3/0&0/4/0&0/4/0\\
		\midrule
		WD~\cite{zhang2019multisource}&4/0/0&3/1/0&3/1/0&3/0/1&2/2/0\\
		\midrule
		MM~\cite{bali2019multifactorial}&4/0/0&3/1/0&3/1/0&2/2/0&2/2/0\\
		\midrule
		RD~\cite{yao2021self}&2/1/1&2/2/0&2/2/0&2/1/1&2/1/1\\
		\midrule
		w/t/l (\%)&69/25/6&63/37/0&56/44/0&44/44/12&38/56/6\\
		\bottomrule
	\end{tabular}
	\label{tab:reliability}
\end{table}
The optimization results of BO-LCB-CKT equipped with the five similarity metrics are provided in the supplementary document due to the page limit, from which we find that the metrics except for WD achieve comparable results to SSRC.
This indicates that the final optimization results of SAS-CKT are not sensitive to the choice of similarity metric.
However, the performance of the five similarity metrics varies when the computational budget is limited.
To investigate, we compare SSRC with the other four metrics under different real function evaluations (FEs) on the MS problems and summarize its wins, ties and losses in Table \ref{tab:reliability}.
It can be seen that SSRC outperforms all the other similarity metrics in cases of limited computational budget and its superiority decreases with the available FEs.
This advantage of SSRC can be attributed to its more reliable similarity quantification as compared to the other similarity metrics in cases of limited target data, enabling it to identify the promising optimized solutions to speed up the target search effectively at the early stage.
This merit is of particularly great significance for EOPs with limited real evaluations available.

\begin{figure}[ht]
	\centering
	\includegraphics[width=3.4in]{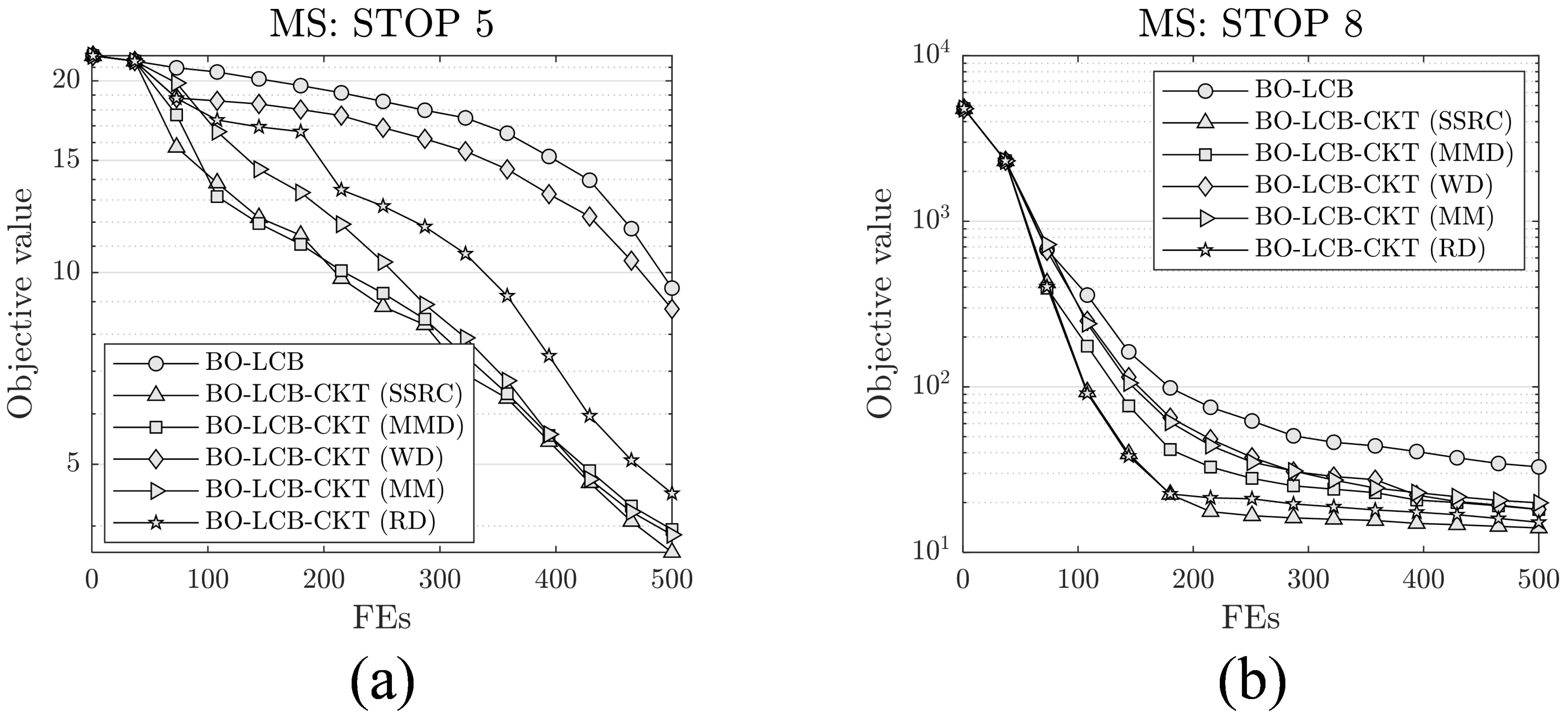}
	\caption{Averaged convergence curves of BO-LCB and BO-LCB-CKT with the five similarity metrics on two problems: (a) STOP 5; (b) STOP 8.}
	\label{fig:reliability_curves}
\end{figure}

As the number of available FEs increases, the other similarity metrics can also identify the promising optimized solutions due to the sufficient target data enabling more accurate similarity measurements.
That explains the comparable results obtained by the five similarity metrics under 500 FEs.
To demonstrate further, we compare the averaged convergence curves of BO-LCB-CKT based on the five similarity metrics in Fig. \ref{fig:reliability_curves}.
It can be seen that SSRC shows the best convergence speedup among the five similarity metrics, especially in cases of limited FEs.

\subsubsection{Comparison with the State-of-the-art} 
\begin{figure}[ht]
	\centering
	\includegraphics[width=3.4in]{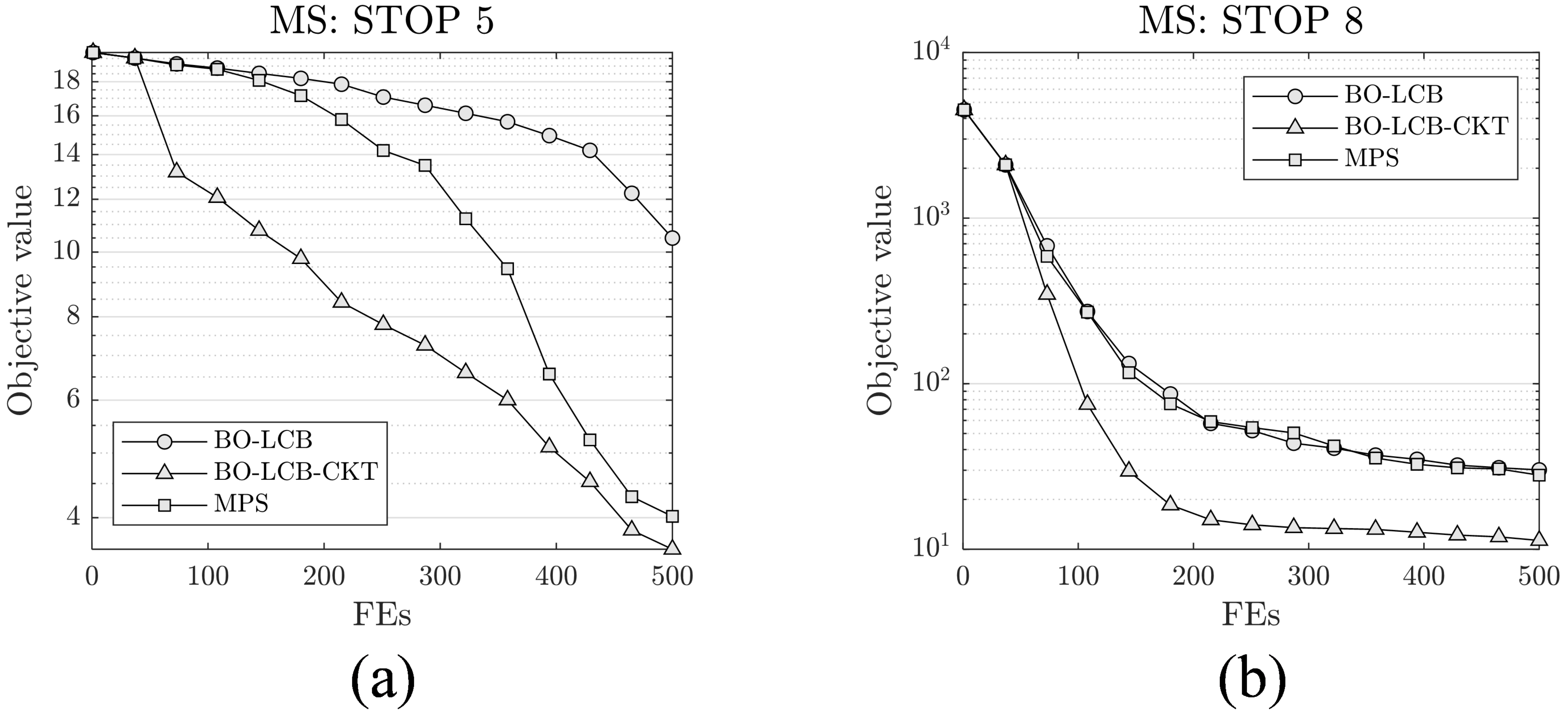}
	\caption{Averaged convergence curves of BO-LCB, BO-LCB-CKT and MPS on two problems: (a) STOP 5; (b) STOP 8.}
	\label{fig:mps_curves}
\end{figure}

To validate the superiority of SAS-CKT over the multi-problem surrogates (MPS) in~\cite{min2019multiproblem}, we compare them on two MS problems in Fig. \ref{fig:mps_curves}.
Specifically, STOP 5 involves the intra-family transfer and its source tasks have the same orders of magnitude to the target task.
By contrast, the inter-family transfer in STOP 8 implies that the source and target tasks in this problem come from different task families and thus may bear heterogeneous orders of magnitude (HOM), which may pose a challenge to the similarity quantification for evaluating the source knowledge.
To make fair comparisons, we employ BO-LCB as the backbone optimizer whose surrogate is consistent with MPS and equalize their surrogate-based evaluations in each acquisition cycle.
The settings of the remaining parameters in MPS are consistent with its original paper~\cite{min2019multiproblem}.

Fig. \ref{fig:mps_curves} shows the averaged convergence curves of BO-LCB, BO-LCB-CKT and MPS on STOP 5 and STOP 8, respectively.
On STOP 5 with the intra-family tasks, both BO-LCB-CKT and MPS achieve effective convergence speedups as compared to the backbone optimizer.
Moreover, we can see that the proposed knowledge competition yields significantly higher convergence gain in comparison with MPS at the early stage of optimization, indicating its superior capability to identify the promising optimized solutions with limited target data.
However, MPS fails to accelerate the target search on STOP 8 while the knowledge competition can still yield the positive convergence gain steadily, as shown in Fig. \ref{fig:mps_curves}(b).
This is because the similarity quantification in SAS-CKT is not sensitive to HOM while the performance of MPS could deteriorate a lot on problems characterized by HOM.
For problems with HOM tasks, the objective value-based similarity quantification in MPS could make it underestimate the promising knowledge hidden in source tasks, e.g., STOP 8 in our experiments.
In summary, the more reliable similarity quantification in SAS-CKT that uses ranks instead of exact objective values accounts for its superiority against MPS in assessing the knowledge from the source tasks.

\subsection{Adaptivity of CKT}
\label{subsection:adaptivity}

To validate the adaptivity of CKT in adapting the optimized solutions from source tasks, we first examine the superiority of the proposed surrogate-based domain adaption (SDA) over a few domain adaptation methods commonly used in transfer optimization~\cite{lin2023ensemble}.
Without loss of generality, we use the interior point algorithm to solve the problem in Eq. \eqref{eq:ada_translation} based on the initial target data.
Subsequently, we discuss the auxiliary role of SDA in SAS-CKT.

\begin{table}[ht]
	\caption{Wins, ties and losses of SDA against the four domain adaptation methods.}
	\centering
	\footnotesize
	\heavyrulewidth=0.12em
	\lightrulewidth=0.1em
	\cmidrulewidth=0.1em
	\setlength\tabcolsep{5.4pt}
	\begin{tabular}{lcccc}
		\toprule
		Method&HS Problems&MS Problems&LS Problems&Summary\\
		\midrule
		TCA~\cite{jiang2017transfer}&1/3/0&4/0/0&1/3/0&6/6/0\\
		\midrule
		AE~\cite{feng2017autoencoding}&2/2/0&3/1/0&1/2/1&6/5/1\\
		\midrule
		AT~\cite{xue2020affine}&2/2/0&4/0/0&1/2/1&7/4/1\\
		\midrule
		SA~\cite{tang2020regularized}&1/3/0&3/1/0&1/3/0&5/7/0\\
		\midrule
		w/t/l (\%)&38/62/0&88/12/0&25/63/12&50/46/4\\
		\bottomrule
	\end{tabular}
	\label{tab:adaptivity}
\end{table}

\subsubsection{Comparison with Other Domain Adaptation Methods}
\begin{figure}[ht]
	\centering
	\includegraphics[width=3.4in]{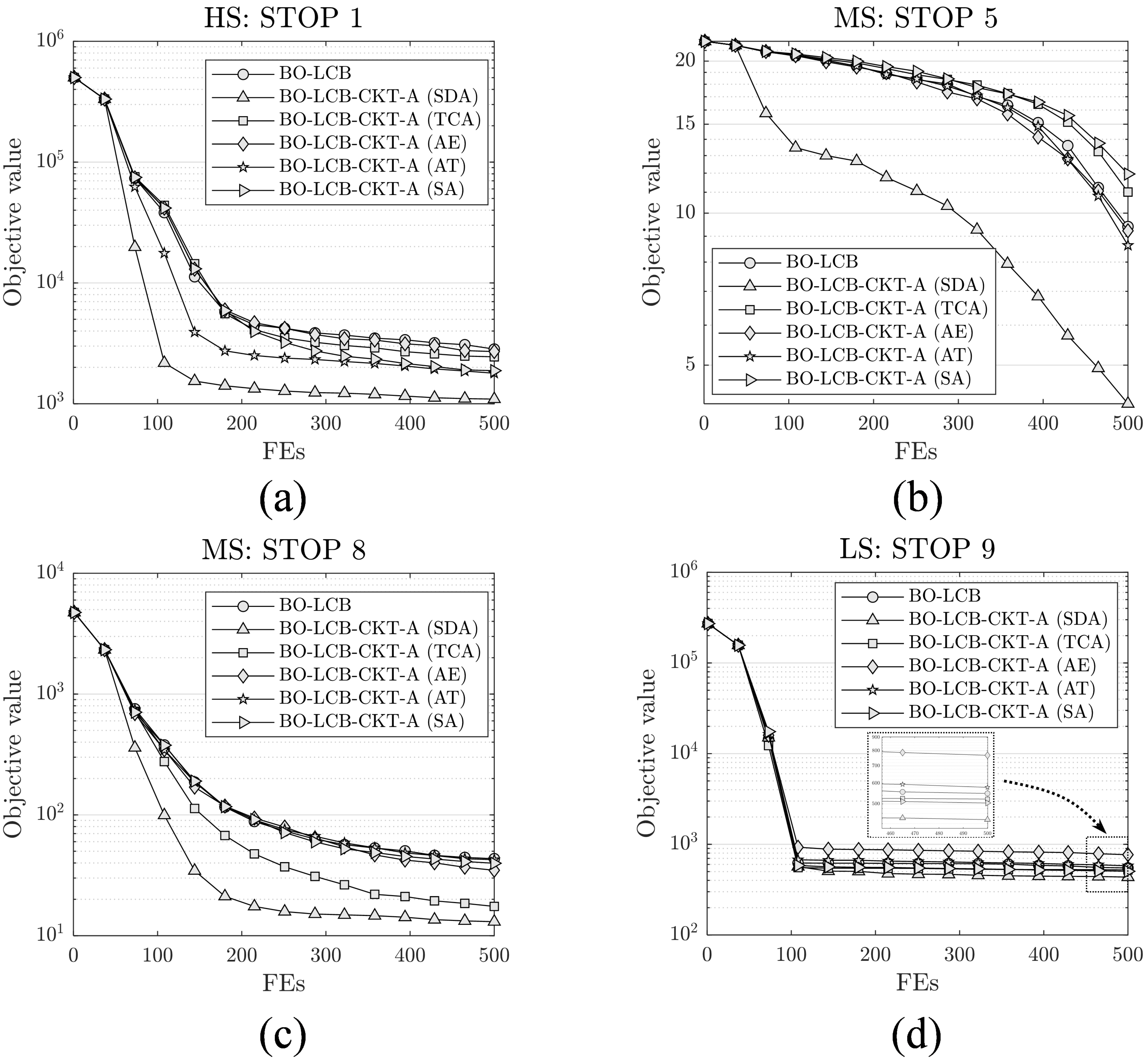}
	\caption{Averaged convergence curves of BO-LCB and BO-LCB-CKT with the five domain adaptation methods on four problems: (a) STOP 1; (b) STOP 5; (c) STOP 8; (d) STOP 9.}
	\label{fig:adaptivity_curves}
\end{figure}

We employ four domain adaptation methods, including the transfer component analysis (TCA)~\cite{jiang2017transfer}, autoencoder (AE)~\cite{feng2017autoencoding}, affine transformation (AT)~\cite{xue2020affine}, and subspace alignment (SA)~\cite{tang2020regularized}, to compare with SDA.
The wins, ties and losses of SDA against the four domain adaptation methods are reported in Table \ref{tab:adaptivity}\footnote{The detailed optimization results are provided in the supplement.}.
We can observe that SDA outperforms all the other domain adaptation methods on both the HS and MS problems.
This advantage of SDA is largely attributed to its mild yet appropriate task adaptation that preserves the high transferability of the promising optimized solutions in the HS or MS problems, enabling it to achieve the best final results among the five domain adaptation methods.
On the LS problems, SDA, AE and AT achieve the best optimization results on STOP 9, STOP 11 and STOP 12, respectively, indicating that these methods exhibit distinct advantages when adapting the optimized solutions of different LS problems.
To illustrate, we compare the averaged convergence curves of BO-LCB-CKT based on the five domain adaptation methods in Fig. \ref{fig:adaptivity_curves}.
It can be seen that SDA shows the best convergence speedups among the five domain adaptation methods.
Moreover, the effective convergence acceleration of SDA on STOP 9 validates the adaptivity of SAS-CKT-A in dealing with LS problems.
It is noted that a more sophisticated surrogate-based domain adaptation technique may yield higher convergence gain on LS problems, but it is still compatible with SAS-CKT, which holds much promise as subjects for future inquiry.

\subsubsection{The Auxiliary Role of SDA}
\begin{figure}[ht]
	\centering
	\includegraphics[width=3.4in]{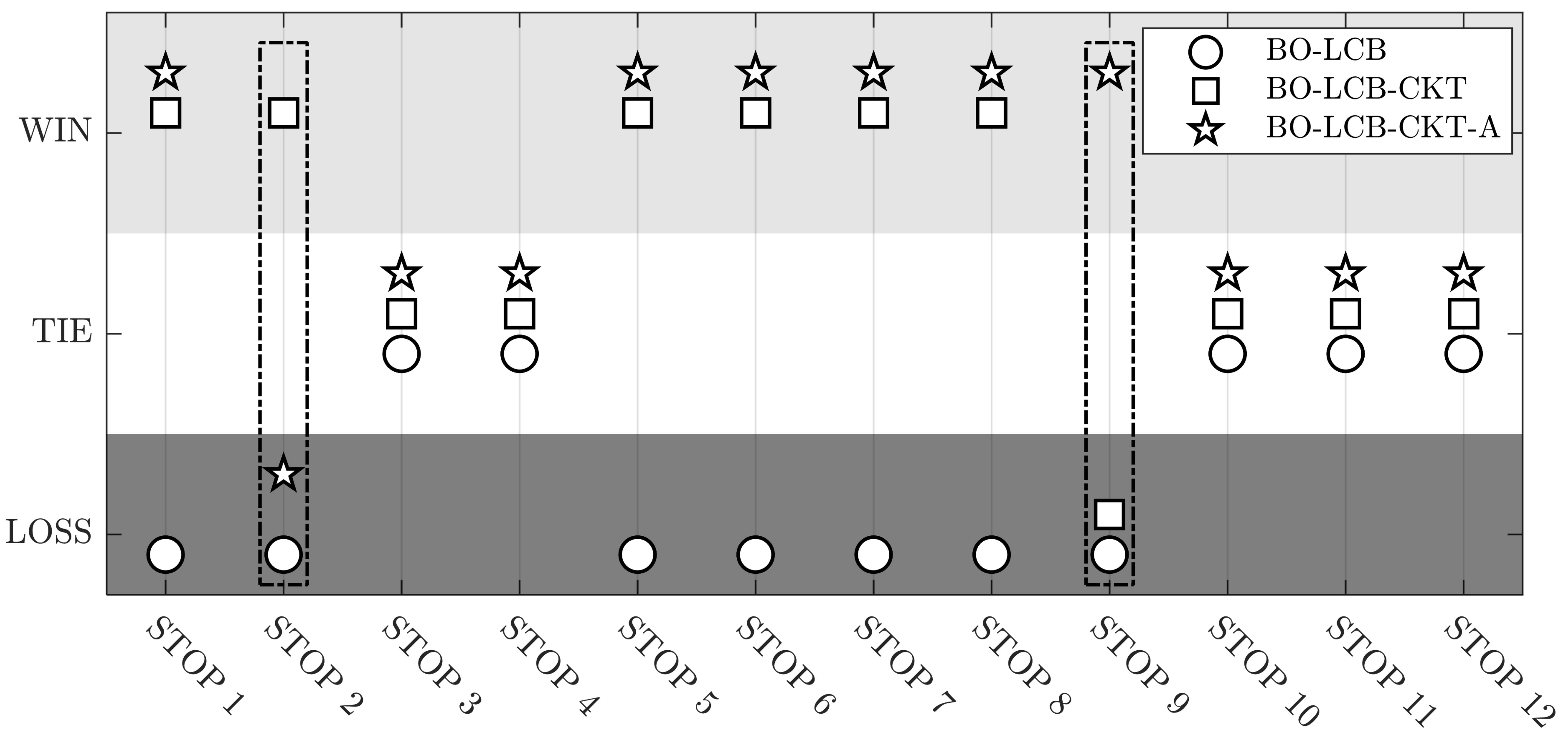}
	\caption{Comparison results of BO-LCB, BO-LCB-CKT and BO-LCB-CKT-A on the 12 test problems.}
	\label{fig:auxiliary}
\end{figure}

To investigate the effect of the task adaptation on the competitive knowledge transfer, we compare BO-LCB with its adaptation-free and adaptation-based knowledge competition methods, as denoted by BO-LCB-CKT and BO-LCB-CKT-A, respectively.
Fig. \ref{fig:auxiliary} demonstrates the comparison results based on the Wilcoxon rank-sum test of their final objective values.
Overall, both CKT and CKT-A are effective in speeding up the target search of BO-LCB, which yield positive performance gain across the majority of the HS and MS problems while exhibiting comparable optimization results to BO-LCB on the remaining problems.
Interestingly, there are two problems wherein CKT and CKT-A show distinguishably dominant convergence speedups, i.e., STOP 2 and STOP 9, as framed by the dotted boxes shown in Fig. \ref{fig:auxiliary}.
This tradeoff between the activation and cancellation of the task adaptation can be attributed to the inductive biases in the adaptation-free and adaptation-based methods.
For the adaptation-free method, i.e., BO-LCB-CKT, its identity transformation (i.e., no adaptation) enables it to focus on measuring the quality of knowledge without paying attention to adapting the knowledge, which fits the nature of HS or MS problems.
By contrast, the learnable transformation in BO-LCB-CKT-A makes it put more emphasis on adapting the knowledge, which fits the nature of LS problems\footnote{More discussions in this respect can be found in Section VI-B of~\cite{xue2023solution}.}.
In summary, the task adaptation can be seen as an auxiliary component of SAS-CKT, whose configuration is supposed to be driven by one's prior knowledge of the similarity relationship of STOPs at hand.
In cases of unknown similarity relationship, both CKT and CKT-A are available for boosting SAS algorithms.

\begin{figure}[ht]
	\centering
	\includegraphics[width=3.4in]{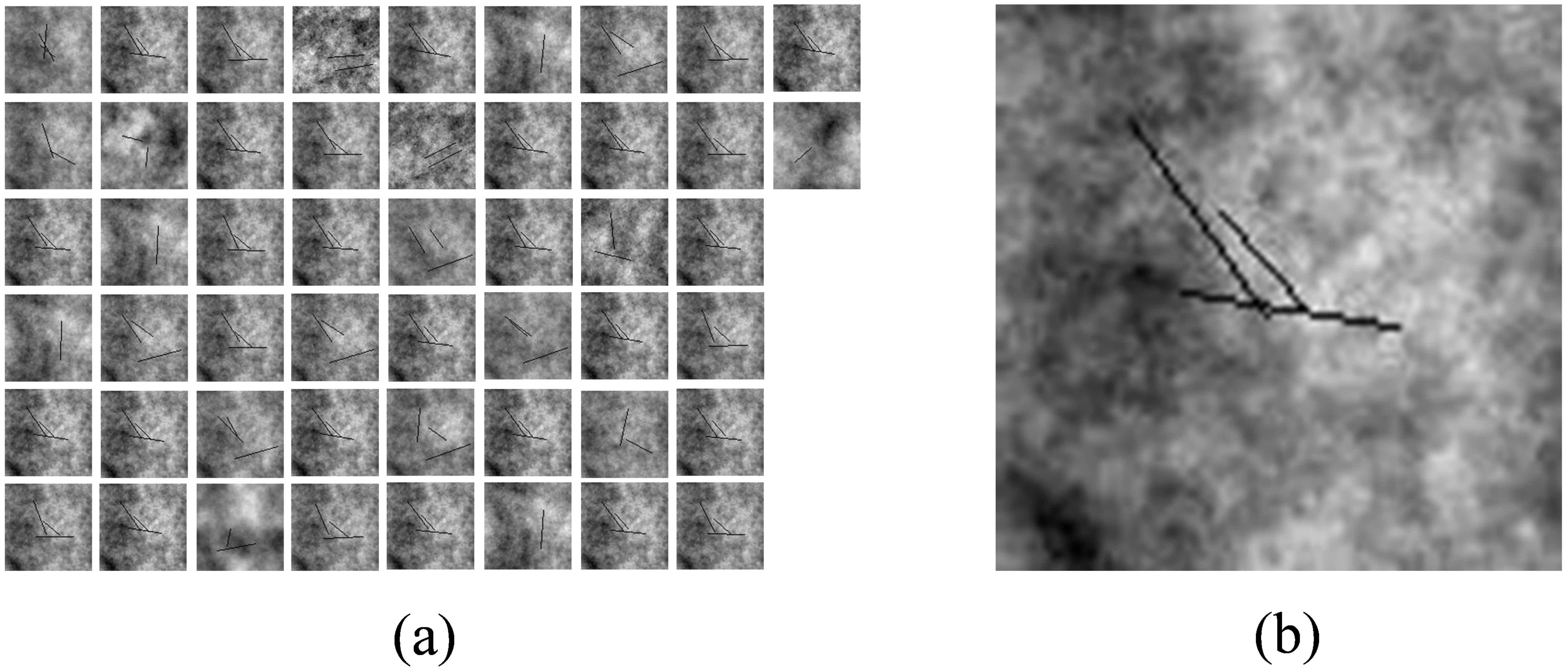}
	\caption{Permeability fields of the source and target tasks: (a) the 50 source tasks; (b) the target task.}
	\label{fig:reservoirs}
\end{figure}

\subsection{A Practical Case Study}

Production optimization in the petroleum industry is a representative EOP that involves high-fidelity numerical simulations, which also necessitates frequent re-optimization in view of newfound reservoirs.
As time goes by, an increasing number of developed reservoirs with available optimization experience accumulate in a database managed by the data processing center, providing an opportunity to the reservoir of interest for achieving better development effect through knowledge transfer.
Thus, we use production optimization to evaluate the efficacy of the proposed method, whose ultimate goal is to search for the optimal production scheme that maximizes the development effect.
The net present value (NPV)~\cite{chen2020global} is commonly used as the objective function for measuring the development effect, which is given by
\begin{equation}
	NPV\left(\boldsymbol{v}\right)=\sum_{t=1}^T Q^t_{po}r_o+Q^t_{pw}r_{wd}+Q^t_{iw}r_{wi},
	\label{eq:NPV}
\end{equation}
where $\boldsymbol{v}$ is the production scheme that consists of production and injection rates, $t$ denotes the $t$-th timestep, $T$ is the total number of timesteps, $Q^t_{po}$ and $Q^t_{pw}$ represent the oil and water outputs of all the producers at the $t$-th timestep, $Q^t_{iw}$ denotes the water input of all the injectors at the $t$-th timestep, $r_o$ is the oil revenue, $r_{wd}$ is the cost of disposing of produced water, $r_{wi}$ denotes the water-injection cost.

In this case study, we have 50 previously-solved production optimization tasks corresponding to 50 developed reservoirs acting as the source tasks and one reservoir model at hand serving as the target task, each of which has a 2D grid pattern of $100\times100\times1$ and is developed by the nine-spot well pattern with five producers and four injectors.
The permeability fields of the source and target tasks are shown in Fig. \ref{fig:reservoirs}.
For the sake of brevity, the remaining reservoir properties are not reported herein but can be found in~\cite{chen2020global}.
The maximum number of simulation calls (i.e., expensive FEs) is set to 200 and all the algorithms are executed for 30 independent runs.

\begin{figure}[ht]
	\centering
	\includegraphics[width=3.4in]{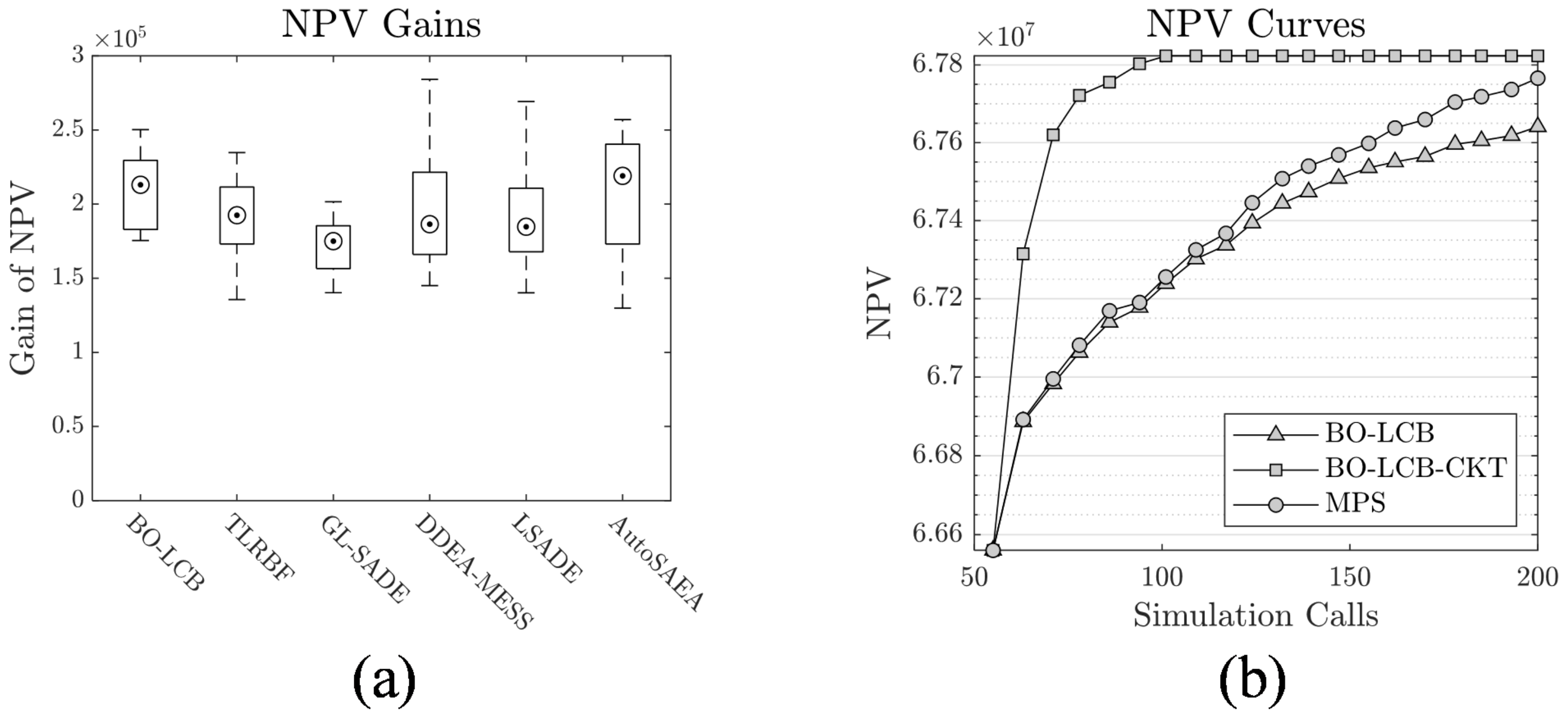}
	\caption{NPV gains obtained by SAS-CKT based on the six SAS optimizers and averaged NPV curves of BO-LCB, BO-LCB-CKT and MPS: (a) the NPV gains; (b) the NPV curves.}
	\label{fig:npvs}
\end{figure}

Fig. \ref{fig:npvs} shows the NPV gains achieved by SAS-CKT across the six backbone optimizers and the averaged convergence curves of BO-LCB, BO-LCB-CKT and MPS over 30 independent runs.
The NPV gain here is defined as the NPV improvement made by an SAS-CKT algorithm as compared with its backbone SAS optimizer.
According to the results in Fig. \ref{fig:npvs}(a), we can see that the NPV gains obtained by the knowledge competition for all the six backbone optimizers are significantly greater than zero, validating its efficacy in boosting different SAS algorithms on the production optimization.
From the averaged NPV curves in Fig. \ref{fig:npvs}(b), we can see that the knowledge competition shows better performance in terms of both convergence speed and solution quality as compared to the backbone BO-LCB and MPS, enabling more efficient decision-making for such computationally expensive production optimization problems.


\section{Conclusions}

\label{section:con}

In this paper, we have developed a novel competitive knowledge transfer (CKT) method for improving surrogate-assisted search (SAS) algorithms, which is capable of leveraging the optimization experience hidden in previously-solved tasks to improve the optimization performance on a task at hand.
The proposed CKT differs from the preceding methods due to its three merits itemized as follows:
\begin{itemize}
	\item [1)] \emph{Portability}: The plug and play nature of CKT enables it to boost different SAS algorithms, making it coherent with the state-of-the-art in surrogate-assisted optimization.
	\item [2)] \emph{Reliability}: The rank-based similarity makes CKT reliable in assessing the knowledge from source tasks, especially in the case of source-target tasks with heterogeneous orders of magnitude.
   \item [3)] \emph{Adaptivity}: The task adaptation enables CKT to adapt the knowledge for higher transferability, making it applicable to problems with heterogeneous tasks.
\end{itemize}

The core of CKT is treating both the optimized solutions from the source tasks and the promising solutions acquired by the target surrogate as task-solving knowledge, enabling us to assess them from a consistent view.
In this way, the most promising solution will be identified for true evaluation to maximize the search speed on the target task.
Specifically, a convergence analogy-based estimation method has been proposed to enable the competition between the optimized solutions from the source tasks and the searched solutions in the target task.
The efficacy of the proposed method has been validated empirically on a series of synthetic benchmarks and a practical application from the petroleum industry.
In addition, we have conducted theoretical analyses on the convergence gain brought by the knowledge competition, including the lower bound of the convergence gain and the important conditions that could yield positive convergence gain.
We anticipate that these analyses can strengthen the theoretical foundation of sequential transfer optimization and help researchers gain a deeper understanding of the underlying conditions contributing to effective knowledge transfer.

Despite the aforementioned merits of the proposed CKT, it exhibits certain limitations, e.g., the zero convergence gain on the LS problems.
On such problems, some novel transferable cues that are not reliant on the similarity in terms of optimum could potentially yield positive convergence gain to the target search, which would be investigated in our future work.
Moreover, we intend to generalize CKT to combat more problem complexities including, but not limited to, objective conflicts in multi-objective problems, high-dimensionality in large-scale problems, and constraint satisfaction in constrained problems.
We also show particular interest in the generalization of CKT to other transfer optimization paradigms, e.g., multi-task and multi-form optimization.

\footnotesize
\bibliography{main}

\newpage
\normalsize

\section*{Supplementary Document}

\setcounter{figure}{0}
\setcounter{table}{0}
\renewcommand\thesection{S-\Roman{section}}
\renewcommand\thefigure{S-\arabic{figure}}
\renewcommand\thedefi{S-\arabic{defi}}
\renewcommand\thelems{S-\arabic{lems}}
\renewcommand\thecoro{S-\arabic{coro}}
\renewcommand\thetable{S-\arabic{table}}

\subsection{Calculations of the Internal Improvement}

When improving a specific SAS optimizer with the proposed knowledge competition method, the calculation of the internal improvement described by Eq. (6) in the main paper depends on the infill criterion used.
To elaborate, we make a number of infill criteria commonly used in surrogate-assisted optimization (as described in~\cite{xie2023surrogate}) fit for the internal improvement and itemize them as follows:

(1) \emph{Predicted Objective Value (POV)}: The internal improvement is estimated based on the predicted objective values of the surrogate model:
\begin{equation}
\mathcal{I}_{\mathrm{pov}}\left(\boldsymbol{x}\right) = \hat{f}\left(\boldsymbol{x}\right),
\label{eq:pov}
\end{equation}
where $\hat{f}\left(\cdot\right)$ denotes the target surrogate model.

(2) \emph{Expected Improvement (EI)}: The internal improvement is calculated based on the expected improvement predicted by the surrogate model:
 \begin{equation}
\mathcal{I}_{\mathrm{ei}}\left(\boldsymbol{x}\right) = \lbrack\hat{f}\left(\boldsymbol{x}\right)-y_{\mathrm{min}}\rbrack\Phi\left(z\right)-\hat{s}\left(\boldsymbol{x}\right)\phi\left(z\right),
\label{eq:ei}
\end{equation}
where $z=\lbrack y_{\mathrm{min}}-\hat{f}\left(\boldsymbol{x}\right)\rbrack/\hat{s}\left(\boldsymbol{x}\right)$, $y_{\mathrm{min}}$ represents the current minimum objective value, $\hat{s}\left(\boldsymbol{x}\right)$ is the prediction uncertainty, $\Phi\left(\cdot\right)$ and $\phi\left(\cdot\right)$ denote the density and cumulative distribution functions of the standard normal distribution, respectively.

(3) \emph{Lower Confidence Bound (LCB)}: The internal improvement is estimated based on the lowed confidence bound:
 \begin{equation}
\mathcal{I}_{\mathrm{lcb}}\left(\boldsymbol{x}\right) = \hat{f}\left(\boldsymbol{x}\right)-w\hat{s}\left(\boldsymbol{x}\right),
\label{eq:lcb}
\end{equation}
where $w$ is a trade-off parameter for balancing the exploitation and exploration, which is set to 2 in this paper.

(4) \emph{L1-Exploration or L1-Exploitation}: For classifier-based surrogate models, we propose to estimate the internal improvement as follows:
\begin{equation}
\Delta_{\mathrm{in}} = \frac{1}{\mid\boldsymbol{y}_{\mathrm{top}}\mid-1}\sum_{i=1}^{\mid\boldsymbol{y}_{\mathrm{top}}\mid-1}y^{i+1}_{\mathrm{top}}-y^i_{\mathrm{top}},
\label{eq:imp_in_classifier}
\end{equation}
where $\boldsymbol{y}_{\mathrm{top}}$ denotes the objective values of the best solutions at the top level, which is sorted in ascending order, $\mid\boldsymbol{y}\mid$ represents the number of elements in $\boldsymbol{y}$.
A new promising solution classified into the top level is supposed to yield an improvement approximated by the averaged difference of the objective values of the top-level solutions.

\subsection{Supporting Lemmas}
This section presents the detailed proofs of the three supporting lemmas, which are essential for analyzing the lower bound of the expected convergence gain in the main paper.

\subsubsection{Approximation of Lipschitz Continuous Functions}
\begin{defi}
Given a continuous function $f\left(\boldsymbol{x}\right):\Omega\to\mathbb{R}$, $f_\epsilon\left(\boldsymbol{x}\right):\Omega\to\mathbb{R}$ is recognized as an $\epsilon$-optimal approximation of $f\left(\boldsymbol{x}\right)$, if for any $\boldsymbol{x}\in\Omega$
\begin{equation}
\mid f\left(\boldsymbol{x}\right)-f_\epsilon\left(\boldsymbol{x}\right)\mid\le\epsilon,
\label{eq:approx}
\end{equation}
where $\epsilon>0$ is the upper bound of approximation error.
\label{defi:approximation}
\end{defi}

\begin{lem}
If the objective function $f\left(\boldsymbol{x}\right):\Omega\to\mathbb{R}$ of a continuous optimization problem satisfies the Lipschitz condition with a dilation constant $K^*$, there exists an $\epsilon$-optimal approximation of $f\left(\boldsymbol{x}\right)$ characterized by a finite number of evaluated solutions.
\end{lem}

\begin{IEEEproof}[Proof] Given that $f\left(\boldsymbol{x}\right):\Omega\to\mathbb{R}$ satisfies the Lipschitz condition with the dilation constant $K^*$ on the metric spaces $\left(\Omega,d_{\boldsymbol{x}}\right)$ and $\left(\mathbb{R},d_{y}\right)$, we have
\begin{equation}
d_{y}\left(f\left(\boldsymbol{x}_1\right),f\left(\boldsymbol{x}_2\right)\right)\le K^* d_{\boldsymbol{x}} \left(\boldsymbol{x}_1, \boldsymbol{x}_2\right),\,\,\,\forall \boldsymbol{x}_1, \boldsymbol{x}_2\in\Omega.
\notag
\end{equation}

Let $d_{y}$ be the Manhattan distance according to Definition \ref{defi:approximation}, then $f\left(\boldsymbol{x}_1\right)$ is an $\epsilon$-optimal approximation of $f\left(\boldsymbol{x}_2\right)$ if
\begin{equation}
d_{\boldsymbol{x}} \left(\boldsymbol{x}_1, \boldsymbol{x}_2\right)\le\frac{\epsilon}{K^*}.
\notag
\end{equation}

Next, we construct a neighborhood for $\forall\boldsymbol{x}\in\Omega$ with a radius $r>0$, which is given by
\begin{equation}
B_r\left(\boldsymbol{x}\right) = \lbrace\boldsymbol{x}'\in\Omega:d_{\boldsymbol{x}}\left(\boldsymbol{x},\boldsymbol{x}'\right)\le r\rbrace.
\notag
\end{equation}

Then, $\forall\boldsymbol{x}'\in B_r\left(\boldsymbol{x}\right)$ is an $\epsilon$-optimal approximation of $f\left(\boldsymbol{x}\right)$ if
\begin{equation}
r\le\frac{\epsilon}{K^*}.
\notag
\end{equation}

Let $d_{\boldsymbol{x}}$ be the Chebyshev distance, then the upper bound of the neighborhood's volume can be estimated as follows: 
\begin{equation}
V_B^{\mathrm{u}}=\left(\frac{2\epsilon}{K^*}\right)^d.
\notag
\end{equation}
where $d$ represents the dimension of $\Omega$. The radius of the neighborhood in this case is $r_{\mathrm{u}}=\epsilon/K^*$.

Without loss of generality, we consider $\Omega$ to be the common space (i.e., $\lbrack0,1\rbrack^d$).
Then, $\Omega$ can be partitioned and the number of neighborhoods required for filling $\Omega$ can be calculated as follows:
\begin{equation}
N_B=\frac{V_\Omega}{V_B^{\mathrm{u}}}=\left(\frac{K^*}{2\epsilon}\right)^d.
\notag
\end{equation}

Suppose the centered solutions $X_c$ in the above $N_B$ neighborhoods are evaluated and their objective values are represented by $\boldsymbol{y}_c$, an $\epsilon$-optimal approximation of $f\left(\boldsymbol{x}\right)$ characterized by this set of evaluated solutions is given by
\begin{equation}
f_\epsilon\left(\boldsymbol{x}\right) = \sum_{i=1}^{N_B}\boldsymbol{1}_{B^i_{r_{\mathrm{u}}}}\left(\boldsymbol{x}\right)\cdot y^i_c,
\label{eq:fapprox}
\end{equation}
where $\boldsymbol{1}_B\left(\boldsymbol{x}\right)=1$ if $\boldsymbol{x}\in B$, and $\boldsymbol{1}_B\left(\boldsymbol{x}\right)=0$ otherwise, $B^i_{r_{\mathrm{u}}}$ denotes the $i$-th neighborhood for filling $\Omega$.

The approximation error between $f_\epsilon\left(\boldsymbol{x}\right)$ and $f\left(\boldsymbol{x}\right)$ for $\forall\boldsymbol{x}\in\Omega$ can be calculated as follows:
\begin{equation}
\begin{split}
\mid f\left(\boldsymbol{x}\right)-f_\epsilon\left(\boldsymbol{x}\right)\mid&=f\left(\boldsymbol{x}\right)-y^i_c,\,\,\,\boldsymbol{x}\in B^i_{r_{\mathrm{u}}}\\
&=f\left(\boldsymbol{x}\right)-f\left(\boldsymbol{x}_c^i\right)\\
&\le K^*d_{\boldsymbol{x}}\left(\boldsymbol{x},\boldsymbol{x}_c^i\right)\\
&\le K^*r_{\mathrm{u}}\\
&=\epsilon.
\end{split}
\notag
\end{equation}

\end{IEEEproof}

\subsubsection{Global Convergence of SAS}
\begin{lems}
Let $\lbrace U_t: t\in T\rbrace$ be a sequence of random variables on a probability space $\left(\Omega,\mathcal{F},P\right)$ with values in a set $E$ of measurable space $\left(E,\mathcal{A}\right)$, then $U_t$ is said to converge completely to 0, if for $\forall \xi>0$
\begin{equation}
\lim_{t\to\infty} \sum_{i=1}^t P\left(\mid U_i\mid>\xi\right)<\infty.
\notag
\end{equation}
and to converge in probability to 0, if for $\forall \xi>0$
\begin{equation}
\lim_{t\to\infty} P\left(\mid U_i\mid>\xi\right)=0.
\notag
\end{equation}
\label{lem:convergence}
\end{lems}

In the context of SAS, we let $b(U_t)=\min_{1\le i\le t}f\left(\boldsymbol{x}^i\right)$ denote the best objective value found so far, where $t$ is represented by the total number of evaluated solutions in the database, $\boldsymbol{x}^i$ denotes the $i$-th evaluated solution. According to Lemma \ref{lem:convergence}, a formal definition about the global convergence of SAS is presented as follows:
\begin{defi}
An SAS optimizer converges to the global optimum of $f\left(\boldsymbol{x}\right)$ if the random sequence $W_t=f\left(\boldsymbol{x}_*\right)-b(U_t)$ converges completely to zero, if for any $\xi>0$
\begin{equation}
\lim_{t\to\infty} \sum_{i=1}^t P\left(\mid f\left(\boldsymbol{x}_*\right)-\min_{1\le i\le t}f\left(\boldsymbol{x}^i\right)\mid>\xi\right)<\infty,
\notag
\end{equation}
where $\boldsymbol{x}_*$ denotes the global optimum.
\end{defi}

Before we prove the global convergence of SAS, let us show an auxiliary convergence result of EA~\cite{rudolph1996convergence}:
\begin{lems}
An EA with an elite selection and a mutation kernel $\mathbb{K}_m\left(\boldsymbol{x},A\right)$ that is strictly bounded from zero for each $\boldsymbol{x}\in E$ and $A_\xi\in\mathcal{A}$ converges to the global optimum of a real-valued function $f\left(\boldsymbol{x}\right):\Omega\to\mathbb{R}$ with $f>-\infty$ defined on an arbitrary space $\Omega$, where $A_\xi=\lbrace\boldsymbol{x}\in E: \mid f\left(\boldsymbol{x}_*\right)-f\left(\boldsymbol{x}\right)\mid\le\xi\rbrace$ with $\forall\xi>0$ denotes a set of $\xi$-optimal states.
\label{lem:ea_convergence}
\end{lems}

Essentially, an SAS optimizer can be seen as a special EA with a single individual and the solution modification scheme being replaced by a specific surrogate-assisted search method.
Thus, Lemma \ref{lem:ea_convergence} is still valid for proving the global convergence of SAS, which is formally given by
\begin{lem}
An SAS optimizer with the database maintenance and the prohibitive evaluation for neighboring solutions converges to the global optimum of a real-valued Lipschitz continuous function $f\left(\boldsymbol{x}\right):\Omega\to\mathbb{R}$ with $f>-\infty$.
\label{lem:sas_convergence}
\end{lem}

\begin{IEEEproof}[Proof] For a real-valued Lipschitz continuous function $f\left(\boldsymbol{x}\right)$ with a dilation constant $K^*$, we will show that the global convergence on its $\epsilon$-optimal approximation function $f_\epsilon\left(\boldsymbol{x}\right)$ is a sufficient condition for the global convergence on $f\left(\boldsymbol{x}\right)$. The proof of the convergence on $f_\epsilon\left(\boldsymbol{x}\right)$ relies on an important corollary of Lemma \ref{lem:ea_convergence}: the global convergence on problems with finite sample spaces can be guaranteed by an EA with an elite selection kernel and a positive mutation kernel. There are three conditions that sufficiently lead to the global convergence:
\begin{itemize}
	\item The finiteness of sample space.
	\item The positiveness of mutation kernel.
	\item The eliteness of selection kernel.
\end{itemize}

Next, we can examine these three conditions separately in SAS.
Firstly, $f_\epsilon\left(\boldsymbol{x}\right)$ can be treated as a discrete problem due to its lookup table-based evaluation in Eq. \eqref{eq:fapprox}. In this sense, the sample space of $f_\epsilon\left(\boldsymbol{x}\right)$ is finite and its size is $\left(K^*/2\epsilon\right)^d$.

Secondly, let us figure out what does the positiveness of mutation kernel imply in the context of SAS.
In EAs, a positive mutation kernel guarantees that $A_\xi$ is reachable in one step from any $\boldsymbol{x}\in E$ at least with probability $\delta>0$.
In SAS, a widely used strategy that prohibits the evaluation of a new solution when it is sufficiently close to the evaluated ones can produce the same effect.
For an $\epsilon$-optimal approximation function $f_\epsilon\left(\boldsymbol{x}\right)$, $\boldsymbol{x}_1$ and $\boldsymbol{x}_2$ are deemed sufficiently close if $d_{\boldsymbol{x}} \left(\boldsymbol{x}_1, \boldsymbol{x}_2\right)\le r_{\mathrm{u}}=\epsilon/K^*$.
If the new solution $\boldsymbol{x}^{t+1}$ is not sufficiently close to every evaluated solution, that is 
\begin{equation}
\boldsymbol{x}^{t+1}\in L\left(P_t\right)=\lbrace\boldsymbol{x}\in E: d_{\boldsymbol{x}}\left(\boldsymbol{x},\boldsymbol{v}\right)> r_{\mathrm{u}},\,\forall\boldsymbol{v}\in P_t\rbrace,
\notag
\end{equation}
where $P_t=\lbrace\boldsymbol{x}^1,\boldsymbol{x}^2,...,\boldsymbol{x}^t\rbrace$ denotes the set of the evaluated solutions, then the new solution will be evaluated and added into the database, otherwise the new solution is regenerated until it is evaluable.
This process can be explicitly described as follows:
\begin{equation}
\boldsymbol{x}^{t+1}=\boldsymbol{x}^{t+1}\cdot\boldsymbol{1}_{L\left(P_t\right)}\left(\boldsymbol{x}^{t+1}\right)+\mathbb{H}_{L\left(P_t\right)}\cdot\boldsymbol{1}_{L^c\left(P_t\right)}\left(\boldsymbol{x}^{t+1}\right),
\notag
\end{equation}
where $L^c\left(P_t\right)=E\backslash L\left(P_t\right)$, $\mathbb{H}_{L\left(P_t\right)}$ can be the reacquisition by SAS for producing an evaluable solution or an additional sampler that samples a solution from $L\left(P_t\right)$.

Thirdly, the eliteness of selection kernel is satisfied by SAS with ease due to its database maintenance.
Since the best solution ever found will always be archived by the database, we denote the state of the best solution as $Q_t=\min_{1\le i\le t}f_\epsilon\left(\boldsymbol{x}^i\right)$.
If the quality of the newly evaluated solution $\boldsymbol{x}^{t+1}$ is not worse than $Q_t$, that is
\begin{equation}
f_\epsilon\left(\boldsymbol{x}^{t+1}\right)\in M\left(Q_t\right)=\lbrace f_\epsilon\left(v\right)\mid v\in E: f_\epsilon\left(v\right)\le Q_t\rbrace,
\notag
\end{equation}
then the state of the best solution will be updated, otherwise it remains unchanged.
This process can be formally formulated as follows:
\begin{equation}
Q_{t+1} = Q_{t+1}\cdot \boldsymbol{1}_{M\left(Q_t\right)}\left(Q_{t+1}\right)+Q_{t}\cdot\boldsymbol{1}_{M^c\left(Q_t\right)}\left(Q_{t+1}\right),
\notag
\end{equation}
where $M^c\left(Q_t\right)=\lbrace f_\epsilon\left(v\right)\mid v\in E:f_\epsilon\left(v\right)>Q_t\rbrace$.
It is evident that the eliteness of selection is intrinsically satisfied by SAS, whose Markovian kernel is explicitly given by
\begin{equation}
\begin{split}
\mathbb{K}_s\left(M_{t+1},F_\epsilon;M_t\right)=\,&\boldsymbol{1}_{M\left(Q_t\right)}\left(Q_{t+1}\right)\cdot\boldsymbol{1}_{F_\epsilon}\left(M_{t+1}\right)+\\
&\boldsymbol{1}_{M^c\left(Q_t\right)}\left(Q_{t+1}\right)\cdot\boldsymbol{1}_{F_\epsilon}\left(M_{t}\right),
\end{split}
\notag
\end{equation}
where $F_\epsilon$ denotes the objective space of $f_\epsilon\left(\boldsymbol{x}\right)$.

After $N_B=\left(K^*/2\epsilon\right)^d$ evaluations, the sample space of $f_\epsilon\left(\boldsymbol{x}\right)$ will be exhausted.
Therefore, for any $\xi>0$ and $t> N_B$, we have
\begin{equation}
\begin{split}
P\left(\mid f^*_{\epsilon}-Q_t\mid>\xi\right)&=P\left(\mid f^*_{\epsilon}-\min_{1\le\ i\le N_B}f_{\epsilon}\left(\boldsymbol{x}^i\right)\mid>\xi\right)\\
&=P\left(\mid f^*_{\epsilon}-f^*_{\epsilon}\mid>\xi\right)\\
&=0.
\end{split}
\notag
\end{equation}

Then, we obtain
\begin{equation}
\lim_{t\to\infty} \sum_{i=1}^t P\left(\mid f_{\epsilon}^*-\min_{1\le i\le t}f_{\epsilon}\left(\boldsymbol{x}^i\right)\mid>\xi\right)<\infty.
\notag
\end{equation}
The global convergence on $f_\epsilon\left(\boldsymbol{x}\right)$ is proved: an SAS optimizer can converge to $f_\epsilon^*$ using $\left(K^*/2\epsilon\right)^d$ evaluations at most.

Let the global optimum of $f\left(\boldsymbol{x}\right)$ be denoted by $\boldsymbol{x}_*$.
For any $t> N_B$, we have
\begin{equation}
\mid W_t\mid=\mid f\left(\boldsymbol{x}_*\right)-\min_{1\le\ i\le N_B}f_{\epsilon}\left(\boldsymbol{x}^i\right)\mid.
\notag
\end{equation}
Let $\boldsymbol{x}_{\mathrm{min}}$ be the optimal solution of $f_{\epsilon}\left(\boldsymbol{x}\right)$. If $\boldsymbol{x}_{\mathrm{min}}$ and $\boldsymbol{x}_*$ are in the same neighborhood, we have
\begin{equation}
\mid W_t\mid=\mid f\left(\boldsymbol{x}_*\right)-f_{\epsilon}\left(\boldsymbol{x}_{\mathrm{min}}\right)\mid\le\epsilon.
\notag
\end{equation}
When $\boldsymbol{x}_{\mathrm{min}}$ and $\boldsymbol{x}_*$ are in different neighborhoods, we have
\begin{equation}
\begin{split}
\mid W_t\mid&=\mid f\left(\boldsymbol{x}_*\right)-f_{\epsilon}\left(\boldsymbol{x}_{\mathrm{min}}\right)\mid\\
&\le\mid f\left(\boldsymbol{x}_*\right)-f\left(\boldsymbol{x}_{\mathrm{min}}\right)\mid+\mid f_{\epsilon}\left(\boldsymbol{x}_{\mathrm{min}}\right)-f\left(\boldsymbol{x}_{\mathrm{min}}\right)\mid\\
&\le2\epsilon+\epsilon\\
&=3\epsilon.
\notag
\end{split}
\end{equation}
The above results indicate that, for any $\xi>0$, the global convergence of an SAS optimizer on an $\epsilon$-optimal approximation $f_{\epsilon}\left(\boldsymbol{x}\right)$ of $f\left(\boldsymbol{x}\right)$ with $\epsilon\le\xi/3$ can guarantee the global convergence on $f\left(\boldsymbol{x}\right)$.
For any $t>N_B$, we have
\begin{equation}
P\left(\mid W_t\mid>\xi\right)=P\left(\max\lbrace\epsilon,3\epsilon\rbrace)>\xi\right)=0.
\notag
\end{equation}
Thus, we have
\begin{equation}
\lim_{t\to\infty} \sum_{i=1}^t P\left(\mid f\left(\boldsymbol{x}_*\right)-\min_{1\le i\le t}f\left(\boldsymbol{x}^i\right)\mid>\xi\right)<\infty.
\notag
\end{equation}
\end{IEEEproof}

\subsubsection{Sufficiency of Similarity for Optimum Equivalence}
\begin{defi}Given two real-valued functions $f^a\left(\boldsymbol{x}\right)$ and $f^b\left(\boldsymbol{x}\right)$, their optimal solutions $\boldsymbol{x}^a_*$ and $\boldsymbol{x}^b_*$ are defined as $\eta$-equivalent if for any $\eta>0$
\begin{equation}
\begin{cases}
\mid f^a\left(\boldsymbol{x}^a_*\right)-f^a\left(\boldsymbol{x}^b_*\right)\mid\le\eta,\\
\mid f^b\left(\boldsymbol{x}^b_*\right)-f^b\left(\boldsymbol{x}^a_*\right)\mid\le\eta.
\end{cases}
\label{eq:opt_equ}
\end{equation}
\end{defi}

\begin{lem}
The optimal solutions of two Lipschitz continuous functions $f^a\left(\boldsymbol{x}\right)$ and $f^b\left(\boldsymbol{x}\right)$ are $\eta$-equivalent if the Spearman's rank correlation coefficient between the neighborhoods' objective values of their $\epsilon$-optimal approximations $f^a_\epsilon\left(\boldsymbol{x}\right)$ and $f^b_\epsilon\left(\boldsymbol{x}\right)$ with $\epsilon\le\eta/6$ is equal to 1.
\label{lem:sim_oe}
\end{lem}
\begin{IEEEproof}[Proof] Suppose the objective values of the $\epsilon$-optimal approximations' neighborhoods are denoted by $\boldsymbol{y}^a$ and $\boldsymbol{y}^b$, then a Spearman's rank correlation value of 1 between $\boldsymbol{y}^a$ and $\boldsymbol{y}^b$ implies that
\begin{equation}
\mathcal{R}\left(y^a_i\right)=\mathcal{R}\left(y^b_i\right),\,\,\,\forall i\in\lbrace1,2,...,N_B\rbrace.
\notag
\end{equation}
where $\mathcal{R}\left(y_i\right)$ represents the rank of $y_i$.
There exists a solution $\boldsymbol{x}_{\mathrm{min}}$ that minimizes both $f^a_\epsilon\left(\boldsymbol{x}\right)$ and $f^b_\epsilon\left(\boldsymbol{x}\right)$, which is given by
\begin{equation}
\boldsymbol{x}_{\mathrm{min}} = \lbrace\boldsymbol{x}\in X^a:\mathcal{R}_{\boldsymbol{x}}=1\rbrace\cap\lbrace\boldsymbol{x}\in X^b:\mathcal{R}_{\boldsymbol{x}}=1\rbrace,
\notag
\end{equation}
where $X^a$ and $X^b$ denote the solutions in the $N_B$ neighborhoods of $f^a_\epsilon\left(\boldsymbol{x}\right)$ and $f^b_\epsilon\left(\boldsymbol{x}\right)$, respectively, $\mathcal{R}_{\boldsymbol{x}}$ denote $\mathcal{R}\left(y\mid\boldsymbol{x}\right)$.

Let the global optima of $f^a\left(\boldsymbol{x}\right)$ and $f^b\left(\boldsymbol{x}\right)$ be denoted by $\boldsymbol{x}_*^a$ and $\boldsymbol{x}_*^b$, respectively.
Next, we will prove the optimum equivalence of $\boldsymbol{x}_*^b$ for $f^a\left(\boldsymbol{x}\right)$ only, since the other one can be proved with ease by interchanging the arguments.
For brevity, let $L =\mid f^a\left(\boldsymbol{x}^a_*\right)-f^a\left(\boldsymbol{x}^b_*\right)\mid$.
When $\boldsymbol{x}_*^a$, $\boldsymbol{x}_*^b$ and $\boldsymbol{x}_{\mathrm{min}}$ are in the same neighborhood, we have $L\le\epsilon<\eta$.
When $\boldsymbol{x}_*^a$ and $\boldsymbol{x}_{\mathrm{min}}$ are in the same neighborhood but $\boldsymbol{x}_*^b$ is in a different one, we have
\begin{equation}
\begin{split}
L&\le\mid f^a\left(\boldsymbol{x}_*^a\right)-f_{\epsilon}^a\left(\boldsymbol{x}_{\mathrm{min}}\right)\mid+\mid f^a\left(\boldsymbol{x}_*^b\right)-f_{\epsilon}^a\left(\boldsymbol{x}_{\mathrm{min}}\right)\mid\\
&\le\mid f^a\left(\boldsymbol{x}_*^a\right)-f^a\left(\boldsymbol{x}_{\mathrm{min}}\right)\mid+\mid f_{\epsilon}^a\left(\boldsymbol{x}_{\mathrm{min}}\right)-f^a\left(\boldsymbol{x}_{\mathrm{min}}\right)\mid\\
&\,\,\,\,\,\,+\mid f^a\left(\boldsymbol{x}_*^b\right)-f^a\left(\boldsymbol{x}_{\mathrm{min}}\right)\mid+\mid f_{\epsilon}^a\left(\boldsymbol{x}_{\mathrm{min}}\right)-f^a\left(\boldsymbol{x}_{\mathrm{min}}\right)\mid\\
&\le\epsilon+\epsilon+2\epsilon+\epsilon\\
&=5\epsilon.
\notag
\end{split}
\end{equation}
The proof for the case of $\boldsymbol{x}_*^b$ and $\boldsymbol{x}_{\mathrm{min}}$ in the same neighborhood while $\boldsymbol{x}_*^a$ in a different one can be done in the same way by swapping the arguments.
When $\boldsymbol{x}_*^a$, $\boldsymbol{x}_*^b$ and $\boldsymbol{x}_{\mathrm{min}}$ are in different neighborhoods, we have
\begin{equation}
\begin{split}
L&\le\mid f^a\left(\boldsymbol{x}_*^a\right)-f_{\epsilon}^a\left(\boldsymbol{x}_{\mathrm{min}}\right)\mid+\mid f^a\left(\boldsymbol{x}_*^b\right)-f_{\epsilon}^a\left(\boldsymbol{x}_{\mathrm{min}}\right)\mid\\
&\le2\epsilon+\epsilon+2\epsilon+\epsilon\\
&=6\epsilon.
\notag
\end{split}
\end{equation}
Thus, we have $L\le\max\lbrace\epsilon,5\epsilon,6\epsilon\rbrace=6\epsilon\le\eta$.

\end{IEEEproof}

Lemma \ref{lem:sim_oe} shows that the optimum equivalence error between two Lipschitz functions can be bounded by their $\epsilon$-optimal approximations. Notably, the exact equivalence requires an infinite number of neighborhoods, that is
\begin{equation}
\lim_{\eta\to0} N_B = \lim_{\epsilon\to0} \left(\frac{K^*}{2\epsilon}\right)^d=\infty.
\notag
\end{equation}

According to Lemma \ref{lem:sas_convergence} and Lemma \ref{lem:sim_oe}, we have an important corollary, which is given by
\begin{coro} Given source and target tasks whose objective functions $f^s\left(\boldsymbol{x}\right)$ and $f^t\left(\boldsymbol{x}\right)$ are Lipschitz continuous, transfer of globally converged source solution searched by SAS will make the target task converge to its global optimum immediately if the rank correlation value between the neighborhood solutions of $f_\epsilon^s\left(\boldsymbol{x}\right)$ and $f_\epsilon^t\left(\boldsymbol{x}\right)$ is 1 for $\forall\epsilon>0$.
\end{coro}
	\begin{IEEEproof}[Proof] Let the best source solution searched by SAS on $f_{\epsilon}^s\left(\boldsymbol{x}\right)$ be denoted by $\boldsymbol{x}^s_{\mathrm{min}}$, then it is in the neighborhood in where $f_{\epsilon}^t\left(\boldsymbol{x}\right)$ achieves the minimum.
Suppose $\boldsymbol{x}^s_{\mathrm{min}}$ is transferred to the target task at time $t$, we have
\begin{figure}[ht]
	\centering
	\includegraphics[width=3.4in]{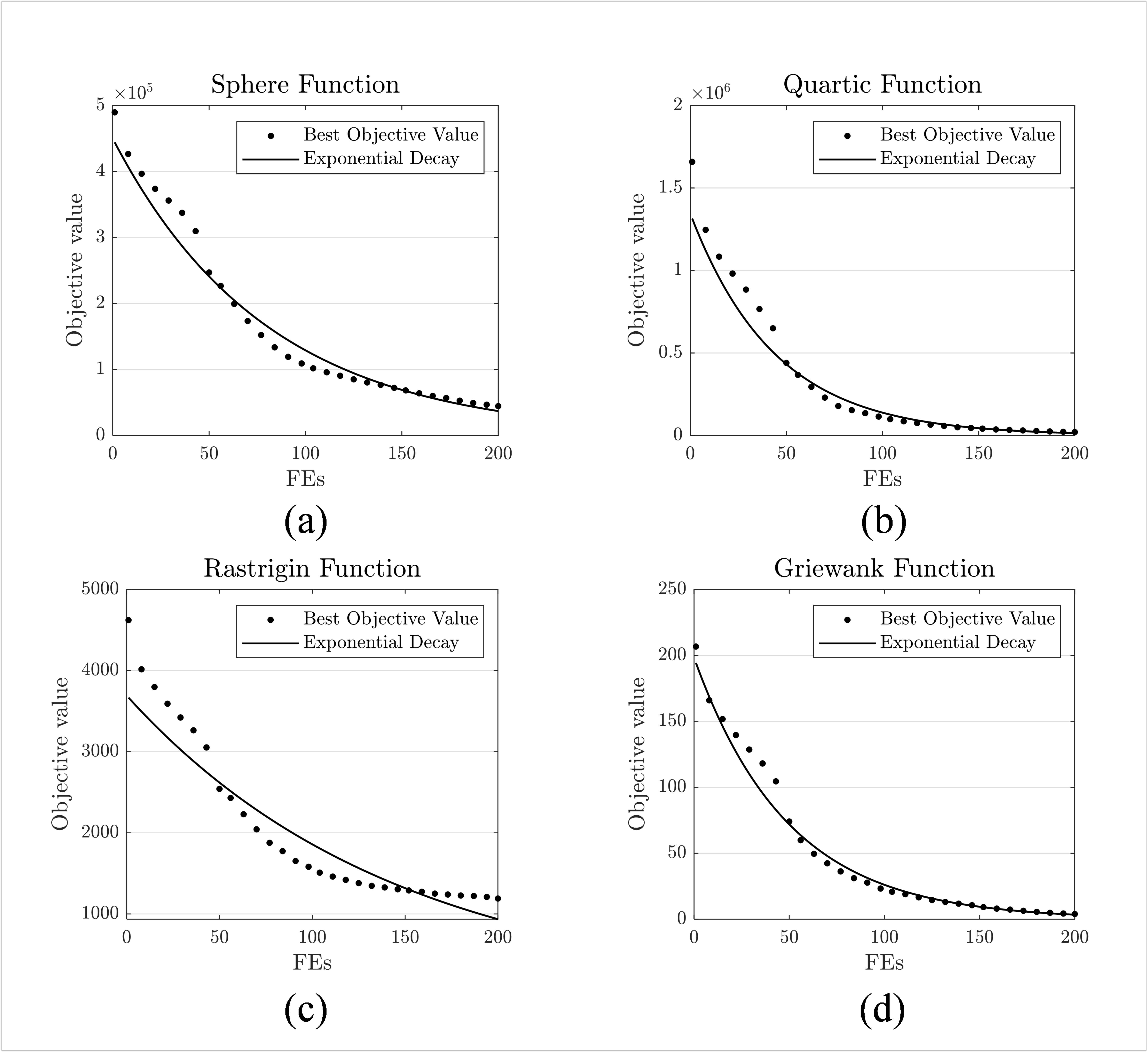}
	\caption{Fitting results of the exponential decay model to the best objective values searched by SAS on four tasks: (a) the Sphere function; (b) the Quartic function; (c) the Rastrigin function; (d) the Griewank function.}
	\label{fig:exps}
\end{figure}
\begin{equation}
\begin{split}
\mid W_t\mid&=\mid f^t\left(\boldsymbol{x}^t_*\right)-f^t\left(\boldsymbol{x}^s_{\mathrm{min}}\right)\mid\\
&\le\mid f^t\left(\boldsymbol{x}^t_*\right)-f^t_{\epsilon}\left(\boldsymbol{x}^s_{\mathrm{min}}\right)\mid+\mid f^t\left(\boldsymbol{x}^s_{\mathrm{min}}\right)-f^t_{\epsilon}\left(\boldsymbol{x}^s_{\mathrm{min}}\right)\mid\\
&\le3\epsilon+\epsilon.
\notag
\end{split}
\end{equation}
For any $\xi\ge\eta\ge4\epsilon>0$, we have $P\left(\mid W_t\mid>\xi\right)=0$, indicating that the global convergence of the target task can be achieved immediately by transferring $\boldsymbol{x}^s_{\mathrm{min}}$.
\end{IEEEproof}

\subsection{Exponential Decay of Best Objective Values}
In this section, we first present the parameter estimation of the exponential decay model.
Then, the empirical evidence for supporting the exponentially decayed best objective values searched by SAS is provided.

\subsubsection{Parameter Estimation}
The exponential decay for modeling the best objective values searched by SAS is given by
\begin{equation}
\gamma_\tau=\gamma_o+\gamma_i e^{-\lambda\tau},
\label{eq:exp}
\end{equation}
where $\gamma_\tau$ denotes the best objective value ever found till time $\tau$, $\gamma_o$ is the optimal objective value, $\gamma_i$ denotes the initial objective value, $\lambda$ represents the decay constant.

In cases of known optimal objective value, i.e., $\gamma_o$, one can set $\gamma_i$ to be the initial objective value and estimate $\lambda$ with the linear least squares.
When $\gamma_o$ is unknown, one can differentiate $\gamma_\tau$ first to eliminate the impact of $\gamma_o$.
Then, with the estimated values of $\gamma_i$ and $\lambda$, one can obtain $\gamma_o$ by subtracting the reconstructed model from the observed data.

\subsubsection{Model Fitting Results}

\begin{table}[ht]
	\caption{Parameter settings of the backbone SAS engines.}
	\centering
	\fontsize{7.5pt}{7.5pt}\selectfont
	\heavyrulewidth=0.12em
	\lightrulewidth=0.1em
	\cmidrulewidth=0.1em
	\setlength\tabcolsep{4pt}
	\begin{tabular}{{llc}}
		\toprule
		Algorithm&Parameter&Value\\
		\midrule
		\multirow{12}*{BO-LCB}&\{surrogate, infill criterion\}&\{GPR, LCB\}\\
		\cmidrule(){2-3}
		&search strategy&iteration\\
		\cmidrule(){2-3}
		&optimizer&evolutionary algorithm (EA$^\dagger$)\\
		\cmidrule(){2-3}
		&$\cdots$initialization&random (50\%) + elite (50\%)\\
		\cmidrule(){2-3}
		&$\cdots$operators&SBX crossover \& poly mutation\\
		\cmidrule(){2-3}
		&$\cdots$\{$p_c,\eta_c,p_m,\eta_m$\}&\{1, 15, $1/d$, 15\}\\
		\cmidrule(){2-3}
		&$\cdots$selector&roulette wheel\\
		\midrule
		\multirow{10}*{TLRBF}&surrogate&RBF\\
		\cmidrule(){2-3}
		&global search: \{$\alpha,m$\}&\{0.4, 200$\times d$\}\\
		\cmidrule(){2-3}
		&subregion search: \{$L_1$, $L_2$\}&\{100, 80\}\\
		\cmidrule(){2-3}
		&local search: \{$k$\}&\{2$\times d$\}\\
		\cmidrule(){2-3}
		&$\cdots$search strategy&iteration\\
		\cmidrule(){2-3}
		&$\cdots$\{opt, infill criterion\}&\{EA$^\dagger$, POV\}\\
		\midrule
		\multirow{12}*{GL-SADE}&\{global model, infill criterion\}&\{RBF, POV\}\\
		\cmidrule(){2-3}
		&$\cdots$search strategy&iteration\\
		\cmidrule(){2-3}
		&$\cdots$optimizer&differential evolution (DE$^\dagger$)\\
		\cmidrule(){2-3}
		&$\cdots\cdots$initialization&elite\\
		\cmidrule(){2-3}
		&$\cdots\cdots$\{operator, $F$, $CR$\}&\{DE/best/1, 0.5, 0.8\}\\
		\cmidrule(){2-3}
		&\{local model, infill criterion\}&\{GPR, LCB\}\\
		\cmidrule(){2-3}
		&$\cdots$\{search strategy, opt\}&\{prescreening, DE$^\dagger$\}\\
		\midrule
		\multirow{12}*{DDEA-MESS}&surrogate&RBF\\
		\cmidrule(){2-3}
		&global search: \{$m$\}&\{300\}\\
		\cmidrule(){2-3}
		&$\cdots$search strategy&prescreening\\
		\cmidrule(){2-3}
		&$\cdots$optimizer&DE with DE/rand/1\\
		\cmidrule(){2-3}
		&local search: \{$\tau$\}&\{25+$d$\}\\
		\cmidrule(){2-3}
		&$\cdots$\{search strategy, opt\}&\{iteration, DE$^\dagger$\}\\
		\cmidrule(){2-3}
		&trust region search: \{$m$\}&\{5$\times d$\}\\
		\midrule
		\multirow{7}*{LSADE}&search strategy&prescreening\\
		\cmidrule(){2-3}
		&global search: \{opt\}&\{DE$^\dagger$\}\\
		\cmidrule(){2-3}
		&Lipschitz search: \{$\alpha$, opt\}&\{0.01, DE$^\dagger$\}\\
		\cmidrule(){2-3}
		&Local search: \{$c$, opt\}&\{3$\times d$, SQP\}\\
		\midrule
		\multirow{7}*{AutoSAEA}&surrogate&\{GPR, RBF, PRS, KNN\}\\
		\cmidrule(){2-3}
		&infill criterion&\{EI, LCB, POV, L1-I, L1-R\}\\
		\cmidrule(){2-3}
		&$\cdots$search strategy&prescreening\\
		\cmidrule(){2-3}
		&$\cdots$optimizer&DE$^\dagger$\\
		\bottomrule
	\end{tabular}
	\label{tab:parameter_settings}
\end{table}

To rationalize the exponential decay in Eq. \eqref{eq:exp} for modeling the best objective values searched by SAS, we examine the goodnesses of fit on four single-objective minimization tasks, including the Sphere, Quartic, Rastrigin, and Griewank functions.
Fig. \ref{fig:exps} shows the observed best objective values searched by SAS and the estimated exponential decay models on the four functions.
It can be seen that the exponential decay can fit the optimization trace of SAS very well, indicating its capability of modeling the best objective values found by SAS.


\begin{figure}[ht]
	\centering
	\includegraphics[width=3.4in]{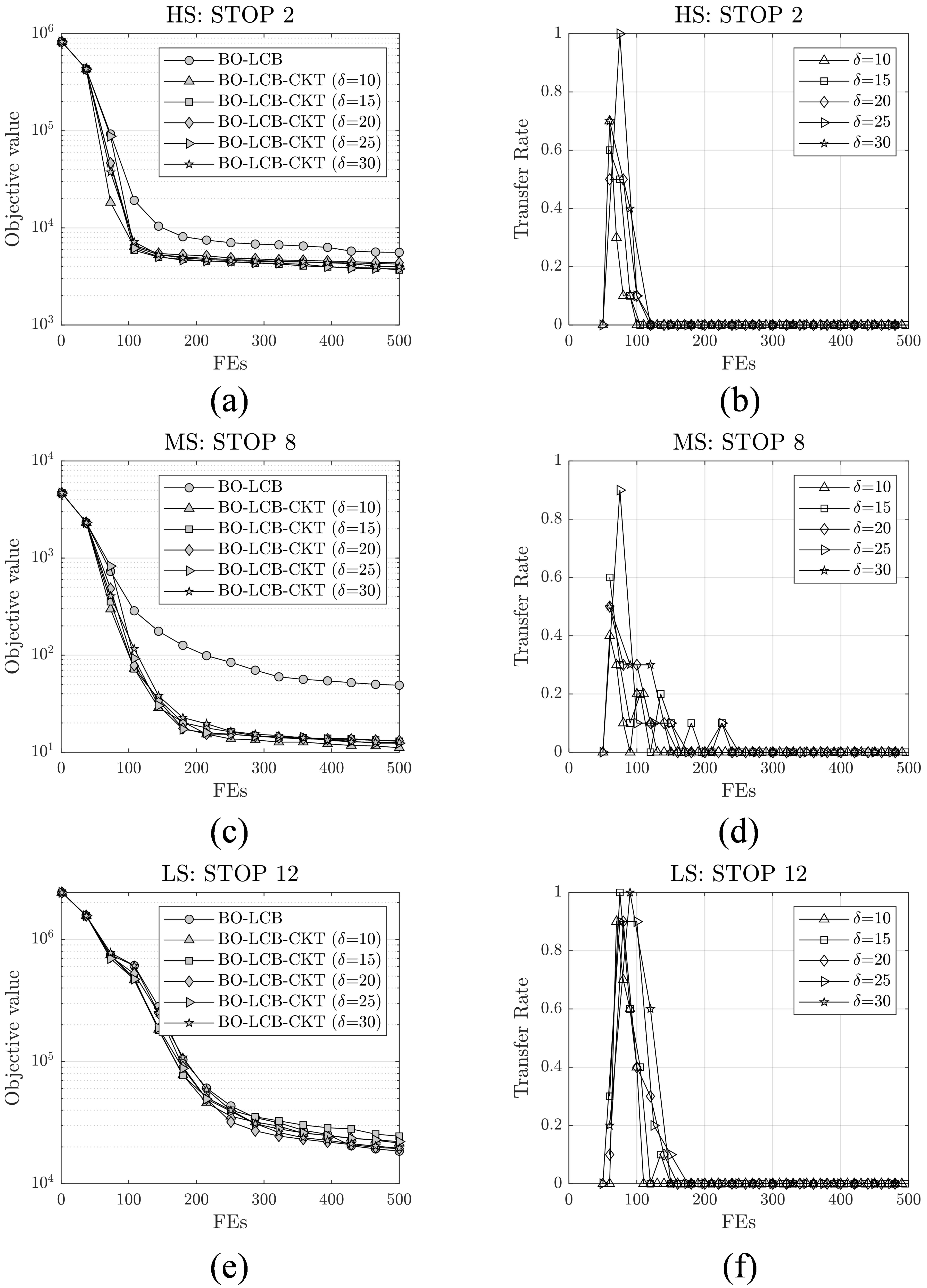}
	\caption{Averaged convergence curves and transfer rates of SAS-CKT with different transfer intervals on three problems: (a) \& (b) STOP 2; (c) \& (d) STOP 8; (e) \& (f) STOP 12.}
	\label{fig:sensitivity}
\end{figure}

\subsection{Detailed Parameter Settings}

Table \ref{tab:parameter_settings} presents the detailed parameter settings of the six backbone SAS optimizers used in the experiments of the main paper.
The settings not reported in the table are consistent with their original papers.


\subsection{Sensitivity Analysis on the Transfer Interval}

To investigate the sensitivity of SAS-CKT to the transfer interval, as denoted by $\delta$, we employ BO-LCB as the backbone optimizer and configure its knowledge competition-based algorithms with different transfer intervals, including 10, 15, 20, 25 and 30.
Based on the Wilcoxon rank-sum test of the final objective values, we find that the knowledge competition-based BO-LCB algorithms with the five transfer intervals show comparable results across the 12 test problems, indicating that the performance of BO-LCB-CKT is not sensitive to the settings of the transfer interval.
To demonstrate further, Fig. \ref{fig:sensitivity} shows the averaged convergence curves and transfer rates of BO-LCB-CKT with the five transfer intervals against BO-LCB over 30 independent runs on three problems.
From the averaged convergence curves, we can see that the five BO-LCB-CKT algorithms with the different transfer intervals exhibit comparable convergence performance on the three problems.
The insensitivity of BO-LCB-CKT to the transfer interval is largely attributed to the proposed knowledge competition that can release the potential of the optimized solutions from source tasks appropriately.
With the estimated external improvements, it is capable of promoting (and curbing) the solution transfer adaptively.
From the averaged transfer rate curves in Fig. \ref{fig:sensitivity}, we can see that BO-LCB-CKT suppresses the knowledge transfer after the utilization of a few optimized solutions regardless of the transfer interval.
At the later stage of optimization, BO-LCB-CKT algorithms based on different transfer intervals are essentially identical as their decisions on whether to transfer the optimized solutions are very similar.
This explains the insensitivity of BO-LCB-CKT to the transfer interval.


\subsection{Detailed Experimental Results}

Table \ref{tab:portability_results} provides the optimization results of the proposed SAS-CKT equipped with the six backbone SAS optimizers across the 12 test problems over 30 independent runs.
The best performing algorithms are highlighted based on the Wilcoxon rank-sum test with Holm $p$-value correction ($\alpha=0.05$).

Table \ref{tab:reliability_results} and Table \ref{tab:adaptivity_results} present the optimization results of BO-LCB-CKT with the comparisons of similarity metrics and domain adaptation methods, respectively, on the 12 test problems over 30 independent runs.
The best performing algorithms are highlighted based on the significance test results.

\begin{table*}[ht]
	\caption{Optimization results of SAS-CKT against SAS with the six backbone optimizers on the 12 test problems. Mean and standard deviation (i.e., mean$\pm$std) are based on 30 independent runs. The winner in each comparison between SAS and SAS-CKT is highlighted based on the Wilcoxon rank-sum test with Holm $p$-value correction ($\alpha=0.05$).}
	\centering
	\heavyrulewidth=0.12em
	\lightrulewidth=0.08em
	\cmidrulewidth=0.05em
	\scriptsize
	\setlength\tabcolsep{4.4pt}
\begin{tabular}{*{15}{lcccccccccccc}}
	\toprule
	\multirow{3}*{Optimizer}&\multirow{3}*{Transfer}&\multirow{3}*{Stat.}&\multicolumn{4}{c}{HS Problems}&\multicolumn{4}{c}{MS Problems}&\multicolumn{4}{c}{LS Problems}\\
	\cmidrule(l){4-7}\cmidrule(l){8-11}\cmidrule(l){12-15}
	&&&STOP 1&STOP 2&STOP 3&STOP 4&STOP 5&STOP 6&STOP 7&STOP 8&STOP 9&STOP 10&STOP 11&STOP 12\\
	\midrule
	\multirow{6}*{BO-LCB}&\multirow{3}*{/}&mean&2.60e+03&5.04e+03&4.73e+02&1.39e+03&9.38e+00&4.17e+02&1.49e+00&3.24e+01&6.56e+02&2.12e+02&2.06e+01&1.73e+04\\
	\cmidrule(){3-15}	&&std&1.05e+03&1.70e+03&3.26e+01&5.52e+02&3.69e+00&2.66e+01&1.72e-01&3.75e+01&2.31e+02&2.02e+01&2.31e-01&5.98e+03\\
	\cmidrule(){2-15}
	&\multirow{3}*{CKT}&mean&\cellcolor[rgb]{0.74,0.74,0.74}1.07e+03&\cellcolor[rgb]{0.74,0.74,0.74}3.01e+03&4.72e+02&1.53e+03&\cellcolor[rgb]{0.74,0.74,0.74}3.67e+00&\cellcolor[rgb]{0.74,0.74,0.74}4.01e+02&\cellcolor[rgb]{0.74,0.74,0.74}1.27e+00&\cellcolor[rgb]{0.74,0.74,0.74}1.32e+01&6.63e+02&2.10e+02&2.06e+01&2.00e+04\\
	\cmidrule(){3-15}	&&std&1.06e+02&1.77e+03&4.02e+01&8.22e+02&1.52e+00&3.27e+01&4.46e-02&3.73e+00&3.15e+02&1.41e+01&1.82e-01&6.79e+03\\
	\midrule
	\multirow{6}*{TLRBF}&\multirow{3}*{/}&mean&3.49e+04&1.17e+04&2.80e+23&1.10e+04&2.06e+01&1.35e+03&4.27e+00&6.84e+02&2.74e+03&5.92e+02&2.11e+01&1.72e+05\\
	\cmidrule(){3-15}	&&std&1.08e+04&3.84e+03&1.53e+24&6.25e+03&2.31e-01&1.35e+02&1.49e+00&1.49e+02&8.54e+02&9.51e+01&8.05e-02&4.40e+04\\
	\cmidrule(){2-15}
	&\multirow{3}*{CKT}&mean&\cellcolor[rgb]{0.74,0.74,0.74}1.29e+03&\cellcolor[rgb]{0.74,0.74,0.74}3.00e+03&3.95e+20&\cellcolor[rgb]{0.74,0.74,0.74}1.28e+02&\cellcolor[rgb]{0.74,0.74,0.74}8.94e+00&\cellcolor[rgb]{0.74,0.74,0.74}4.71e+02&\cellcolor[rgb]{0.74,0.74,0.74}1.40e+00&\cellcolor[rgb]{0.74,0.74,0.74}4.41e+01&3.11e+03&6.13e+02&\cellcolor[rgb]{0.74,0.74,0.74}2.10e+01&1.86e+05\\
	\cmidrule(){3-15}	&&std&1.81e+01&6.98e+02&1.50e+21&9.86e+01&1.21e+00&1.24e+02&2.26e-16&1.35e+01&1.37e+03&9.11e+01&1.11e-01&4.02e+04\\
	\midrule
	\multirow{6}*{GL-SADE}&\multirow{3}*{/}&mean&1.54e+01&5.79e-02&4.83e+02&1.56e+03&1.52e+01&6.63e+02&9.65e-01&2.47e+02&9.57e-03&3.20e+02&1.90e+01&1.02e+02\\
	\cmidrule(){3-15}	&&std&5.38e+01&2.56e-02&5.84e+01&1.05e+03&4.04e+00&6.76e+01&6.22e-02&1.16e+02&5.28e-03&3.11e+01&2.98e+00&9.75e+01\\
	\cmidrule(){2-15}
	&\multirow{3}*{CKT}&mean&1.21e+01&6.18e-02&4.83e+02&1.34e+03&1.44e+01&6.44e+02&9.87e-01&\cellcolor[rgb]{0.74,0.74,0.74}1.01e+02&9.46e-03&3.24e+02&1.92e+01&1.13e+02\\
	\cmidrule(){3-15}	&&std&3.46e+01&3.21e-02&4.86e+01&1.12e+03&5.00e+00&8.17e+01&3.17e-02&1.23e+02&4.62e-03&2.93e+01&2.63e+00&1.18e+02\\
	\midrule
	\multirow{6}*{\makecell[l]{DDEA-\\MESS}}&\multirow{3}*{/}&mean&4.18e-04&5.90e-01&2.17e+18&1.73e+01&3.56e+00&6.79e+02&1.85e-01&2.68e+02&6.09e-05&3.03e+02&1.65e+01&2.72e+02\\
	\cmidrule(){3-15}	&&std&5.53e-04&1.05e+00&1.02e+19&1.23e+01&8.77e-01&8.74e+01&2.12e-01&1.57e+02&1.85e-05&5.77e+01&5.68e+00&2.07e+02\\
	\cmidrule(){2-15}
	&\multirow{3}*{CKT}&mean&2.82e-04&3.95e-01&\cellcolor[rgb]{0.74,0.74,0.74}7.90e+17&\cellcolor[rgb]{0.74,0.74,0.74}5.86e+00&\cellcolor[rgb]{0.74,0.74,0.74}2.73e+00&6.66e+02&1.66e-01&\cellcolor[rgb]{0.74,0.74,0.74}3.89e+01&5.29e-05&3.06e+02&1.80e+01&3.09e+02\\
	\cmidrule(){3-15}	&&std&2.00e-04&4.94e-01&4.32e+18&7.27e+00&1.90e+00&9.29e+01&1.73e-01&6.62e+01&1.64e-05&4.49e+01&4.64e+00&1.58e+02\\
	\midrule
	\multirow{6}*{LSADE}&\multirow{3}*{/}&mean&1.42e-04&1.16e+00&1.76e+23&4.32e+02&6.05e+00&6.65e+02&8.28e-01&2.42e+02&4.85e-05&2.83e+02&1.79e+01&9.43e+02\\
	\cmidrule(){3-15}	&&std&3.72e-05&8.89e-01&7.26e+23&2.72e+02&4.17e+00&9.83e+01&1.09e-01&1.17e+02&1.45e-05&2.95e+01&4.14e+00&4.49e+02\\
	\cmidrule(){2-15}
	&\multirow{3}*{CKT}&mean&\cellcolor[rgb]{0.74,0.74,0.74}1.04e-04&\cellcolor[rgb]{0.74,0.74,0.74}4.38e-01&2.29e+22&\cellcolor[rgb]{0.74,0.74,0.74}3.34e+02&\cellcolor[rgb]{0.74,0.74,0.74}4.13e+00&\cellcolor[rgb]{0.74,0.74,0.74}5.79e+02&8.28e-01&\cellcolor[rgb]{0.74,0.74,0.74}1.49e+01&4.87e-05&2.81e+02&\cellcolor[rgb]{0.74,0.74,0.74}1.58e+01&8.91e+02\\
	\cmidrule(){3-15}	&&std&2.96e-05&2.65e-01&1.25e+23&3.27e+02&4.80e-01&1.29e+02&8.72e-02&1.90e+01&1.66e-05&2.65e+01&4.43e+00&3.90e+02\\
	\midrule
	\multirow{6}*{AutoSAEA}&\multirow{3}*{/}&mean&3.54e+03&1.20e+01&1.10e+15&5.14e+02&6.40e+00&7.42e+02&3.70e-01&1.85e+02&4.27e-01&3.04e+02&2.03e+01&2.32e+04\\
	\cmidrule(){3-15}	&&std&1.62e+03&2.67e+01&6.03e+15&4.04e+02&4.92e+00&1.17e+02&2.95e-01&1.42e+02&1.43e+00&4.26e+01&8.31e-01&6.85e+03\\
	\cmidrule(){2-15}
	&\multirow{3}*{CKT}&mean&\cellcolor[rgb]{0.74,0.74,0.74}2.76e+02&\cellcolor[rgb]{0.74,0.74,0.74}3.57e+00&1.13e+12&\cellcolor[rgb]{0.74,0.74,0.74}1.27e+01&\cellcolor[rgb]{0.74,0.74,0.74}2.73e+00&\cellcolor[rgb]{0.74,0.74,0.74}4.32e+02&\cellcolor[rgb]{0.74,0.74,0.74}2.63e-01&\cellcolor[rgb]{0.74,0.74,0.74}9.18e+00&3.31e-01&2.86e+02&1.96e+01&2.42e+04\\
	\cmidrule(){3-15}	&&std&1.34e+02&1.30e+01&4.47e+12&1.51e+01&7.15e-01&1.69e+02&2.34e-01&6.00e+00&3.22e-01&4.98e+01&1.57e+00&8.11e+03\\
	\bottomrule
	\end{tabular}
	\label{tab:portability_results}
\end{table*}

\begin{table*}[ht]
	\caption{Optimization results of BO-LCB-CKT equipped with the five similarity metrics on the 12 test problems. Mean and standard deviation (i.e., mean$\pm$std) are based on 30 independent runs. The highlighted entries show the best performing algorithm(s) based on the Wilcoxon rank-sum test with Holm $p$-value correction ($\alpha=0.05$).}
	\centering
	\heavyrulewidth=0.12em
	\lightrulewidth=0.08em
	\cmidrulewidth=0.05em
	\scriptsize
	\setlength\tabcolsep{5.5pt}
\begin{tabular}{*{14}{lcccccccccccc}}
	\toprule
	\multirow{3}*{Metric}&\multirow{3}*{Statistics}&\multicolumn{4}{c}{HS Problems}&\multicolumn{4}{c}{MS Problems}&\multicolumn{4}{c}{LS Problems}\\
	\cmidrule(l){3-6}\cmidrule(l){7-10}\cmidrule(l){11-14}
	&&STOP 1&STOP 2&STOP 3&STOP 4&STOP 5&STOP 6&STOP 7&STOP 8&STOP 9&STOP 10&STOP 11&STOP 12\\
	\midrule
	\multirow{3}*{SSRC}&mean&\cellcolor[rgb]{0.74,0.74,0.74}1.05e+03&2.97e+03&4.54e+02&1.53e+03&\cellcolor[rgb]{0.74,0.74,0.74}3.64e+00&\cellcolor[rgb]{0.74,0.74,0.74}3.92e+02&\cellcolor[rgb]{0.74,0.74,0.74}1.28e+00&1.40e+01&6.15e+02&2.12e+02&2.06e+01&1.92e+04\\
	\cmidrule(){2-14}
	&std&1.26e+02&1.59e+03&3.75e+01&5.08e+02&1.29e+00&3.08e+01&4.98e-02&4.49e+00&2.08e+02&1.57e+01&2.14e-01&8.58e+03\\
	\midrule
	\multirow{3}*{MMD}&mean&\cellcolor[rgb]{0.74,0.74,0.74}1.08e+03&3.78e+03&4.70e+02&1.70e+03&\cellcolor[rgb]{0.74,0.74,0.74}3.95e+00&\cellcolor[rgb]{0.74,0.74,0.74}4.06e+02&\cellcolor[rgb]{0.74,0.74,0.74}1.25e+00&1.82e+01&6.85e+02&2.14e+02&2.06e+01&1.88e+04\\
	\cmidrule(){2-14}	&std&1.18e+02&2.03e+03&4.04e+01&1.38e+03&1.81e+00&2.68e+01&4.86e-02&1.58e+01&3.17e+02&2.08e+01&1.70e-01&7.13e+03\\
	\midrule
	\multirow{3}*{WD}&mean&2.58e+03&5.29e+03&2.16e+24&2.16e+03&8.77e+00&\cellcolor[rgb]{0.74,0.74,0.74}3.84e+02&1.53e+00&1.82e+01&6.55e+02&2.07e+02&2.07e+01&1.89e+04\\
	\cmidrule(){2-14}	&std&7.84e+02&2.45e+03&1.17e+25&1.93e+03&3.76e+00&2.86e+01&2.70e-01&9.08e+00&3.37e+02&1.40e+01&2.23e-01&6.55e+03\\
	\midrule
	\multirow{3}*{MM}&mean&\cellcolor[rgb]{0.74,0.74,0.74}1.12e+03&\cellcolor[rgb]{0.74,0.74,0.74}1.55e+03&4.67e+02&1.57e+03&\cellcolor[rgb]{0.74,0.74,0.74}3.88e+00&4.27e+02&\cellcolor[rgb]{0.74,0.74,0.74}1.29e+00&1.99e+01&7.27e+02&2.18e+02&2.06e+01&1.85e+04\\
	\cmidrule(){2-14}	&std&1.01e+02&1.53e+03&4.16e+01&6.15e+02&9.35e-01&2.53e+01&4.71e-02&1.02e+01&3.00e+02&1.64e+01&2.36e-01&6.19e+03\\
	\midrule
	\multirow{3}*{RD}&mean&\cellcolor[rgb]{0.74,0.74,0.74}1.10e+03&2.75e+03&4.65e+02&1.59e+03&4.50e+00&\cellcolor[rgb]{0.74,0.74,0.74}4.12e+02&\cellcolor[rgb]{0.74,0.74,0.74}1.24e+00&1.52e+01&6.28e+02&2.10e+02&2.07e+01&1.95e+04\\
	\cmidrule(){2-14}	&std&1.38e+02&1.69e+03&4.06e+01&6.24e+02&1.24e+00&3.43e+01&4.63e-02&6.98e+00&2.69e+02&1.98e+01&1.68e-01&8.63e+03\\
	\bottomrule
	\end{tabular}
	\label{tab:reliability_results}
\end{table*}

\begin{table*}[ht]
	\caption{Optimization results of BO-LCB-CKT equipped with the five domain adaptation methods on the 12 test problems. Mean and standard deviation are based on 30 independent runs. The highlighted entries show the best performing algorithm(s) based on the Wilcoxon rank-sum test with Holm $p$-value correction ($\alpha=0.05$).}
	\centering
	\heavyrulewidth=0.12em
	\lightrulewidth=0.08em
	\cmidrulewidth=0.05em
	\scriptsize
	\setlength\tabcolsep{5.5pt}
\begin{tabular}{*{14}{lcccccccccccc}}
	\toprule
	\multirow{3}*{Adapter}&\multirow{3}*{Statistics}&\multicolumn{4}{c}{HS Problems}&\multicolumn{4}{c}{MS Problems}&\multicolumn{4}{c}{LS Problems}\\
	\cmidrule(l){3-6}\cmidrule(l){7-10}\cmidrule(l){11-14}
	&&STOP 1&STOP 2&STOP 3&STOP 4&STOP 5&STOP 6&STOP 7&STOP 8&STOP 9&STOP 10&STOP 11&STOP 12\\
	\midrule
	\multirow{3}*{SDA}&mean&\cellcolor[rgb]{0.74,0.74,0.74}1.09e+03&4.12e+03&1.15e+21&2.02e+03&\cellcolor[rgb]{0.74,0.74,0.74}4.19e+00&3.98e+02&\cellcolor[rgb]{0.74,0.74,0.74}1.25e+00&\cellcolor[rgb]{0.74,0.74,0.74}1.31e+01&\cellcolor[rgb]{0.74,0.74,0.74}4.35e+02&2.14e+02&2.07e+01&2.10e+04\\
	\cmidrule(){2-14}
	&std&9.81e+01&1.49e+03&6.31e+21&2.93e+03&1.57e+00&3.48e+01&5.21e-02&3.25e+00&1.01e+02&1.88e+01&1.85e-01&7.50e+03\\
	\midrule
	\multirow{3}*{TCA}&mean&2.45e+03&3.71e+03&4.77e+02&1.64e+03&1.10e+01&4.23e+02&1.39e+00&1.75e+01&5.22e+02&2.14e+02&2.06e+01&2.05e+04\\
	\cmidrule(){2-14}	&std&7.40e+02&1.47e+03&4.38e+01&7.03e+02&3.04e+00&3.36e+01&5.94e-02&7.21e+00&1.34e+02&1.40e+01&1.86e-01&7.88e+03\\
	\midrule
	\multirow{3}*{AE}&mean&2.70e+03&5.28e+03&4.54e+02&1.49e+03&9.23e+00&4.12e+02&1.53e+00&3.48e+01&7.68e+02&2.19e+02&\cellcolor[rgb]{0.74,0.74,0.74}2.02e+01&2.06e+04\\
	\cmidrule(){2-14}	&std&9.49e+02&1.64e+03&4.10e+01&6.20e+02&3.31e+00&2.83e+01&2.91e-01&3.92e+01&4.89e+02&2.15e+01&2.96e-01&7.16e+03\\
	\midrule
	\multirow{3}*{AT}&mean&1.79e+03&5.21e+03&4.55e+02&1.83e+03&8.63e+00&4.23e+02&1.36e+00&4.26e+01&5.79e+02&2.19e+02&2.06e+01&\cellcolor[rgb]{0.74,0.74,0.74}1.56e+04\\
	\cmidrule(){2-14}	&std&3.43e+02&1.23e+03&4.85e+01&1.19e+03&2.92e+00&1.98e+01&1.18e-01&5.21e+01&2.16e+02&1.62e+01&1.54e-01&5.57e+03\\
	\midrule
	\multirow{3}*{SA}&mean&1.88e+03&4.32e+03&4.66e+02&1.45e+03&1.20e+01&4.09e+02&1.64e+00&3.98e+01&5.04e+02&2.15e+02&2.06e+01&2.11e+04\\
	\cmidrule(){2-14}	&std&6.45e+02&1.53e+03&3.87e+01&6.46e+02&3.44e+00&2.87e+01&2.25e-01&4.61e+01&1.13e+02&1.88e+01&1.98e-01&6.83e+03\\
	\bottomrule
	\end{tabular}
	\label{tab:adaptivity_results}
\end{table*}

\ifCLASSOPTIONcaptionsoff
  \newpage
\fi

\end{document}